\documentclass[1p]{elsarticle}

\usepackage{lineno,hyperref}
\modulolinenumbers[5]

\usepackage{amsmath,amsbsy}
\usepackage{booktabs}
\usepackage{xcolor}
\usepackage{graphicx}
\graphicspath{{./figures/}}

\usepackage[caption=false,font=rm, labelfont=rm, textfont=rm]{subfig}

\newcommand{\etal}{\textit{et al}.}

\journal{ISPRS Journal of Photogrammetry and Remote Sensing}

\begin{document}

\begin{frontmatter}

\title{Building Segmentation through a Gated Graph Convolutional Neural Network with Deep Structured Feature Embedding}


\author[mymainaddress]{Yilei Shi}
\ead{yilei.shi@tum.de}

\author[mysecondaryaddress]{Qingyu Li}
\ead{qingyu.li@tum.de}

\author[mysecondaryaddress,mythirdaddress]{Xiao Xiang Zhu\corref{mycorrespondingauthor}}
\cortext[mycorrespondingauthor]{Corresponding author}
\ead{xiaoxiang.zhu@dlr.de}
\ead[url]{www.sipeo.bgu.tum.de}

\address[mymainaddress]{Chair of Remote Sensing Technology, Technical University of Munich, 80333 Munich, Germany}
\address[mysecondaryaddress]{Signal Processing in Earth Observation, Technical University of Munich, 80333 Munich, Germany}
\address[mythirdaddress]{Remote Sensing Technology Institute, German Aerospace Center (DLR), Oberpfaffenhofen, 82234 Wessling, Germany}

\begin{abstract}
Automatic building extraction from optical imagery remains a challenge due to, for example, the complexity of building shapes. Semantic segmentation is an efficient approach for this task. The latest development in deep convolutional neural networks (DCNNs) has made accurate pixel-level classification tasks possible. Yet one central issue remains: the precise delineation of boundaries. Deep architectures generally fail to produce fine-grained segmentation with accurate boundaries due to their progressive down-sampling. Hence, we introduce a generic framework to overcome the issue, integrating the graph convolutional network (GCN) and deep structured feature embedding (DSFE) into an end-to-end workflow. Furthermore, instead of using a classic graph convolutional neural network, we propose a gated graph convolutional network, which enables the refinement of weak and coarse semantic predictions to generate sharp borders and fine-grained pixel-level classification. Taking the semantic segmentation of building footprints as a practical example, we compared different feature embedding architectures and graph neural networks. Our proposed framework with the new GCN architecture outperforms state-of-the-art approaches. Although our main task in this work is building footprint extraction, the proposed method can be generally applied to other binary or multi-label segmentation tasks.
\end{abstract}


\end{frontmatter}


\section{Introduction}
Building footprint generation is an active topic in remote sensing field. Recently, it has received considerable attention due to its huge potential in autonomous driving, virtual reality, urban planning, environmental, and demographic applications. Manual extraction of buildings from optical images is time consuming and difficult in large-scale practice. In contrast, semantic segmentation is a comparatively inexpensive and time-saving technique for extracting building footprints. It aims to classify each pixel with a corresponding class. Various semi-automatic and automatic methods \cite{bib:Ok2013Automated} \cite{bib:Xu2018Building} \cite{bib:Bittner2018Building} \cite{bib:Chen2019Aerial} have been developed to improve segmentation accuracy within this method; traditionally, feature extraction and classification are its two main steps. The extraction of such hand-crafted features usually require a strong domain-specific knowledge.

In recent years, the use of deep learning has garnered great success in semantic segmentation. In particular, deep convolutional neural networks (DCNNs) have shown promising results, due to their high capacity for data learning. DCNNs \cite{bib:zhu2017deep} have instigated compelling advancement over traditional semantic segmentation methods.

However, exploiting DCNN for semantic segmentation tasks still raises significant challenges. The convolution layer of a DCNN is a weights sharing architecture, and it has both shift invariant and spatial invariant characteristics. While the invariance is clearly desirable for high-level vision tasks, it may hamper low-level tasks such as pose estimation and semantic segmentation, where precise localization is required rather than abstraction of spatial details. For instance, the coarse segmentation output such as non-sharp boundaries and blob-like shapes is caused by convolution filters with large receptive fields and pooling layers in DCNN. Moreover, DCNN fails to fine local details without the consideration of the interactions between pixels.

To overcome these issues, the probabilistic graph models, such as the conditional random field (CRF) \cite{bib:chen2017} and Markov random field (MRF) \cite{bib:liu2015}, have been introduced to connect with DCNNs at the final layer. To use CRF for semantic segmentation, the main concept is to transform the problem of pixel-wise classification into a problem of probabilistic inference, which assumes similar pixels should have the same labels. This substantially improves the predictions of the pixel-wise labels to generate precise borders and exhaustive segmentation. In \cite{bib:chen2017}, instead of using CRF as the post-processing step, the authors propose an end-to-end architecture that combines the FCN with a fully connected CRF. However, these frameworks have not sufficiently extracted the features from the images. Different level features have different properties for semantic segmentation. Since low-level characteristics are rich with spatial details but lack semantic information and high-level characteristics are conversely, they are naturally complementary. Another issue with CRF is that information propagation is not sufficient.

In this work, we propose a generic framework for semantic segmentation in this work, which integrates deep structured feature embedding and the graph convolutional network. In order to extract more comprehensive and representative features, we exploit deep structured feature embedding techniques to enhance the feature fusion by incorporating multi-level characteristics. Furthermore, we propose a new graph convolutional network, the gated graph convolutional network (GCN). GCN can aggregate the information from neighbor nodes (short range), which allows the model to learn about local structures. A recurrent neural network (RNN) with gated recurrent units (GRUs) has proven successful to model the long-term dependencies in sequential data. Hence, we adopt RNN with GRUs for long-range information propagation. The proposed network integrates the two architectures together, thus taking into account both local and global contextual dependencies. It is useful for semantic segmentation tasks. As a consequence, DSFE-GGCN is a trainable end-to-end framework. We show that joint learning of deep structured feature embedding and GGCN parameters results in considerable performance gains.

\subsection*{\textbf{Contributions}}
The contributions of this work are summarized as follows:
\begin{itemize}
\item A generic framework for semantic segmentation is proposed, which integrates the deep structured feature embedding and a graph convolutional neural network into an end-to-end workflow.
\item We propose a novel network architecture, called a ``gated graph convolutional neural network,'' which combines the RNN with GRUs for long distance information propagation and the GCN for short distance information propagation.
\item An effective four-step preprocessing approach is proposed for data augmentation, especially for medium-resolution satellite imagery.
\item The performance of different DCNNs and the proposed framework is analyzed through a systematic investigation. Our framework with GGCN surpasses the state-of-the-art approaches to building footprint extraction.
\end{itemize}

\section{Related Work}

\subsection{Semantic segmentation with DCNNs}
The fully convolutional network (FCN) was first proposed in \cite{bib:long2015} for the task of semantic segmentation, in which convolutional layers take the place of fully connected layers. FCN makes the training more efficient and the input size of inference arbitrary. A more memory-efficient approach that used an alternative decoder variant, SegNet, was proposed in \cite{bib:badrinarayanan2015}. The stored indices of the max-pooling step in the downsampling path is used by the decoder for the operation of upsampling. Another variant of the encoder-decoder architecture is U-Net \cite{bib:ronneberger2015}. The long skip connections in the network enables the recovery of the downsample-induced information lost in the encoder.

One key issue for fully convolutional neural networks is that the spatial resolution is significantly downsampled, which is caused by the operations, such as strided convolutional layers or pooling layers. In order to overcome the poor localization property, \cite{bib:zheng2015} proposed another approach to improve the spatial resolution, using a probabilistic graph model CRF to achieve fine-grained boundaries. Instead of using CRF as a post-processing step, DeepLab-CRF \cite{bib:chen2017} introduces a fully connected CRF layer, which leads to an end-to-end trainable network.

\subsection{Graph model}
A graph model is a probabilistic model that encodes a distribution based on a graph-based representation. The Markov random field (MRF) is one classic graph model, which uses an undirected graph to describe the joint probability distribution of random variables. It has been applied to many tasks of image processing, including image co-registration, image segmentation, and image super-resolution. MRF takes into account the relationships of the neighbours to infer the maximal possibility of the pixel's label. The conditional random field (CRF) is an extension of MRF, which models the conditional probability distribution instead of the joint probability distribution. CRF as a discriminative model shows a better performance when the samples are limited. The combination of DCNNs and the graph model CRF \cite{bib:zheng2015, bib:chen2017} can produce high-resolution prediction for better segmentation.

Recent work \cite{bib:bruna2013} has extended DCNNs to topologies that differ from the low-dimensional grid structure. Due to significant computational drawbacks, it is impractical for real-world use. Henaff \etal \cite{bib:henaff2015} and Defferrard \etal \cite{bib:defferrard2016} further improve GCN to successfully overcome this issue. The grid-like data can be interpreted as a special type of graph data, where the node is on the grid and the number of neighbours is fixed. In this work, we propose a gated graph convolutional network, which is a trainable inference systems based on GCN and RNN with GRUs.

\subsection{Building footprint extraction}
Building footprint generation is currently exciting a great deal of interest, and is an active field of research in the fields of remote sensing, photogrammetry and computer vision. The established building footprint maps are used in many important applications to analyze the process of urbanization, such as urban growth and sustainable urban development.

In \cite{bib:yuan2018}, the authors propose a multi-stage ConvNet with an upsampling operation of bilinear interpolation. The trained model achieves a superior performance on very-high-resolution aerial imagery. Recently, an end-to-end trainable active contour model (ACM) was developed for building instance extraction \cite{bib:marcos2018}, which learns ACM parameterizations using a DCNN. In \cite{bib:huang2019}, a residual refinement network was proposed to extract the building footprint using aerial images and LiDAR point clouds. In \cite{bib:shi2018}, the authors exploit the improved conditional Wasserstein generative adversarial network to generate the building footprint automatically. Recent work \cite{bib:wang2017} has shown that most of the tasks, such as building segmentation, building height estimation, and building contour extraction, are still difficult for modern convolutional networks. In this work, we show a significant performance improvement in building footprint extraction by using our proposed novel framework.

\section{Methodology}
\begin{figure*}
  \centering
  \includegraphics[width=\textwidth]{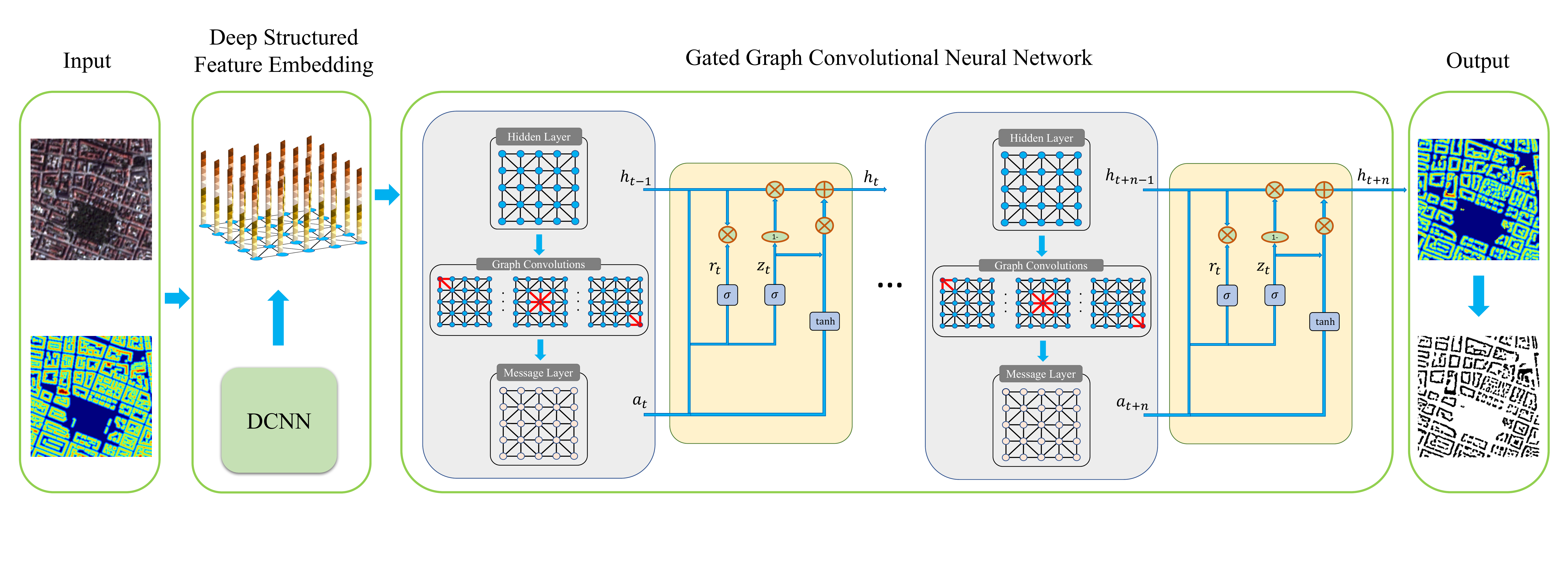}
  \caption{An illustration of the proposed DSFE-GGCN framework. The initial hidden representation of the corresponding node is taken from the feature vectors in the DSFE step. For a certain time step $t$, the messages from the neighbourhoods of the node are aggregated by using GCN. After that, the hidden state of the next time step $t+1$ is updated by gated recurrent units, which use the hidden state $\mathbf{h}_i^{t}$ and the message $\mathbf{a}_i^{t+1}$ at time step $t$ as input. After final timestep $t+n$, a negative log-likelihood loss function is computed and the whole DSFE-GGCN model is updated using back-propagation.}
  \label{fig:DSFE_GGCN_workflow}
\end{figure*}

The details of the DSFE-GGCN framework are introduced in this section. The workflow of the proposed method is shown in Fig. \ref{fig:DSFE_GGCN_workflow}. An image can be generalized as a graph, whose nodes are on the two-dimensional grid. Each pixel represents a node. The embedding vectors can be computed initially from node inputs, e.g., node type embeddings, and then propagated on the graph to aggregate information from the local neighborhood.

\subsection{Deep structured feature embedding}

Deep embedding methods typically map images into an embedding space, where their distances preserve the relative similarity. In general,
the representations of the data can be learned by graph embedding techniques \cite{bib:yan2007}, which take into account the relationships of the data. In addition, data from different sources, such as images, point clouds, and social media data, can be transformed into feature space, which can be further used for segmentation or other tasks. In this study, the data source is only imagery. Hence, we exploit a more efficient approach for feature embedding that uses DCNNs as feature extractor.

However, the resolution of the later layers in the neural network is extremely downsampled, a phenomenon that is caused by strided convolution, max-polling, or other operations. Several methods have been introduced to decipher precise information from the downsampled feature maps. One common approach is to utilize interpolation techniques \cite{bib:badrinarayanan2015}, which is both computationally cheap and memory-saving. An alternative is deconvolution, in which recorded indices of the polling operation are used to retrieve information from the feature maps \cite{bib:noh2015}. Recently, long skip connections between the contracting and expanding paths were introduced to retrieve detailed spatial information from the high-level feature layers \cite{bib:ronneberger2015}. In combination with DenseNet block \cite{bib:huang2017}, FC-DenseNets was proposed in \cite{bib:jegou2017}, where the upsampling path was composed of deconvolution, unpooling, and long skip connections. Consequently, all the feature maps from deconvolution, unpooling, or skip-connections are exploited for the computation in the upsampling path of the dense blocks. Moreover, recent work shows that multiple DCNN features extracted from different networks can be complementary, and could be fused to improve segmentation accuracy. However, the method for fusing multiple features is still an open problem that needs systematic investigation.

As mentioned above, low-level features yield better representation of localization and high-level features can give more comprehensive semantics. Therefore, in this work, we concatenate different level features progressively in order to propagate information about localization, semantics, and other properties through graph convolutional neural networks.

\subsection{Gated graph convolutional neural network}

An undirected and connected graphs $\mathcal{G} = (\mathcal{V}, \mathcal{E})$ consists of a set of nodes $\mathcal{V}$ and edges $\mathcal{E}$. The unnormalized graph Laplacian matrix $\mathbf{L}$ is defined as:
\begin{equation}
 \mathbf{L} = \mathbf{D} - \mathbf{A},
\end{equation}
where $\mathbf{A}$ is the adjacency matrix representing the topology of $\mathcal{G}$, and $\mathbf{D}$ is the degree matrix, which is calculated by $D_{ii} = \sum_j A_{ij}$. The properties of the graph Laplacian $\mathbf{L}$ are symmetric, positive, and semi-defined; therefore the eigenvalue decomposition can be expressed as:
\begin{equation}
  \mathbf{L} = \boldsymbol{\Phi} \boldsymbol{\Lambda} \boldsymbol{\Phi}^{T},
\end{equation}
where $\boldsymbol{\Phi} = \left( \phi_1, \phi_2, ..., \phi_n \right)$ are the orthonormal eigenvectors, known as the graph Fourier modes, and $\boldsymbol{\Lambda} = \mathrm{diag} \left(\lambda_1, \lambda_2, ..., \lambda_n \right)$ are the eigenvalues of $\mathbf{L}$, which is a non-negative diagonal matrix. Assuming a signal $\mathbf{f}$ on the graph nodes $\mathcal{V}$, its graph Fourier transform can be formulated as $\hat{\mathbf{f}} = \boldsymbol{\Phi}^T \mathbf{f}$. If $\mathbf{g}$ is a filter, the convolution of $\mathbf{f}$ and $\mathbf{g}$ is written as
\begin{equation}
  \mathbf{g} * \mathbf{f} = \boldsymbol{\Phi} \left( \left( \boldsymbol{\Phi}^T \mathbf{g}\right) \circ \left( \boldsymbol{\Phi}^T \mathbf{f}\right) \right) = \boldsymbol{\Phi} \hat{\mathbf{g}} \boldsymbol{\Phi}^T \mathbf{f},
\end{equation}
where $\hat{\mathbf{g}}$ is the spectral representation of the filter. Rather than computing the Fourier transform $\hat{\mathbf{g}}$, the filter coefficients can be parameterized as $\hat{\mathbf{g}} = \sum \limits_{k=0}^r \alpha_k \beta_k$, ask shown in \cite{bib:henaff2015}. With the polynomial parametrization of the filter, the spectral filter is exactly localized in space. Moreover, the learning complexity is $\mathcal{O}(r)$, the filter support size, and the same complexity as classical DCNNs.

In order to avoid explicit multiplication in the spectral domain, alternatively, the spectral representation $\hat{\mathbf{g}}$ of the filter $\mathbf{g}$ can be approximated by a Chebyshev polynomial expansion $g(\boldsymbol{\Lambda})$, which is formulated as:
\begin{equation}
  g(\boldsymbol{\Lambda}) = \sum \limits_{k=0}^r \alpha_k T_k(\tilde{\boldsymbol{\Lambda}}),
\end{equation}
where $T_k(\tilde{\boldsymbol{\Lambda}})$ is the Chebyshev polynomials. The graph convolution can be defined as:
\begin{equation}
  \mathbf{g} * \mathbf{f} = \sum \limits_{k=0}^r \alpha_k T_k(\tilde{\mathbf{L}})\mathbf{f},
\end{equation}
where $\tilde{\mathbf{L}} = 2/\lambda_{max} \cdot \mathbf{L} - \mathbf{I}$, and $\lambda_{max}$ is the maximal eigenvector.
In \cite{bib:kipf2016}, the authors further simplify the Chebyshev framework, setting $r = 1$ and assuming $\lambda_{max} \approx 2$, allowing them to redefine a single convolutional layer as simply:
\begin{equation}
  \mathbf{H}_i^{r} = \sigma_r \left( \tilde{\mathbf{D}}^{-1/2} \tilde{\mathbf{A}} \tilde{\mathbf{D}}^{-1/2} \mathbf{W} \mathbf{H}_i^{r-1} \right),
\end{equation}
where $\mathbf{H}$ is the hidden layer. By taking into account the self-connections, the original adjacency matrix of the graph $\mathcal{G}$ is transformed to $\tilde{\mathbf{A}} = \mathbf{A} + \mathbf{I}$, where $\mathbf{I} $ is the identity matrix. $\mathbf{W}$ is the trainable weight matrix and the new degree matrix $\tilde{\mathbf{D}}$ can be calculated by $\tilde{D}_{ii} = \sum_j \tilde{A}_{ij}$. The function $\sigma_r(\cdot)$ denotes a nonlinear activation function. This simplified form improves computational performance on larger graphs and predictive performance on small training sets.

%

\subsubsection*{\textbf{Propagation model}}

The propagation process can be formulated as:
\begin{eqnarray}
  \mathbf{a}_i^{t} &= \mathcal{M}\left(\mathbf{h}_j^{t-1} | j \in \mathcal{V}_i\right), \\
  \mathbf{h}_i^{t} &= \mathcal{F}\left(\mathbf{h}_i^{t-1},  \mathbf{a}_i^{t}\right),
\end{eqnarray}
where $\mathbf{a}_i^{t}$ is the message layer at time step $t$, which represents the messages propagated from its neighbours $\mathcal{V}_i$ to the node $i$. The message layer $\mathbf{a}_i^{t}$ at time step $t$ serves as input to update the hidden layer with function $\mathcal{F}$. Our proposed method is to use GCN as the message function, which makes it easy for the propagation model to learn to propagate the node embeddings for node $i$ to all nodes reachable from $i$. We adopt gating techniques to surpass GCN performance, because its own memory can be maintained and the valuable information from neighbours can be gathered with its aid.

The unrolled propagation model at timestep $t$ can be written as:
\begin{eqnarray}
  \mathbf{a}_i^{t} =& \sigma_r \left( \tilde{\mathbf{D}}^{-1/2} \tilde{\mathbf{A}} \tilde{\mathbf{D}}^{-1/2} \mathbf{W} \mathbf{h}_i^{t-1} \right),\\
  \mathbf{r}_i^{t} =& \sigma_s \left( \mathbf{W}_r \mathbf{h}_i^{t-1} + \mathbf{U}_r \mathbf{a}_i^{t}\right), \\
  \mathbf{z}_i^{t} =& \sigma_s \left( \mathbf{W}_z \mathbf{h}_i^{t-1} + \mathbf{U}_z \mathbf{a}_i^{t}\right), \\
  \tilde{\mathbf{h}}_i^{t} =& \tanh \left( \mathbf{W}\left(\mathbf{r}_i^{t} \circ \mathbf{h}_i^{t-1} \right) + \mathbf{U} \mathbf{a}_i^{t} \right), \\
  \mathbf{h}_i^{t} =& \left( 1 - \mathbf{z}_i^{t}\right) \circ \mathbf{h}_i^{t-1} + \mathbf{z}_i^{t} \circ \tilde{\mathbf{h}}_i^{t},
\end{eqnarray}
where $\mathbf{r}$ and $\mathbf{z}$ are the reset and update gates, and $\mathbf{W}_r, \mathbf{W}_z, \mathbf{U}_r, \mathbf{U}_z$ are learnable weights for different gates. The function $\sigma_r$ is the ReLU function, $\sigma_s$ is the logistic sigmoid function, and $\circ$ is interior product. The initial hidden representation of the corresponding node is taken from the feature vectors of the DSFE step. For a certain time step $t$, the messages from the neighbourhoods of the node are aggregated by using GCN. After that, the hidden state of next time step $t+1$ is updated by gated recurrent units, which use the hidden state $\mathbf{h}_i^{t}$ and the message $\mathbf{a}_i^{t+1}$ at time step $t$ as input. With the help of the reset gate and the update gate in GRU \cite{bib:cho2014}, the node can maintain its own memory and extract useful information from incoming messages. Along with the increase of the time step, it is capable of capturing the long range dependencies, which has been difficult to model in vanilla GCN.

\subsubsection*{\textbf{Prediction model}}
The node classification is defined as:
\begin{equation}
  p = \mathrm{softmax} \left( \mathbf{h}_i^{t+n} \right),
\end{equation}
Since we have transferred the binary semantic segmentation problem to the multi-label pixel labeling task, a softmax with negative log-likelihood loss function is used to predict the probability of each node.

\section{Experiments}
\label{section4}
\subsection{Datasets}
In this work, we use Planetscope satellite imagery \cite{bib:planet} with three channels (R, G, B) at a 3 m spatial resolution. The imagery is acquired by Doves, which can provide complete coverage of Earth once a day. The study sites cover four cities: (1) Munich, Germany; (2) Rome, Italy; (3) Paris, France; and (4) Zurich, Switzerland. The corresponding building footprint layer is downloaded from OpenStreetMap (OSM) \cite{bib:osm}. The images are cropped with a patch size of $\textrm{64} \times \textrm{64}$. The overlap of each patch has 19 pixels in one direction. At the end, 48,000 sample patches are generated. The training data has 80\% of the patches and the testing data has 20\% of the patches. The training and testing data is spatially separated.

\subsection{Preprocessing}
\begin{figure}[!ht]
  \centering
  \includegraphics[width=0.95\textwidth]{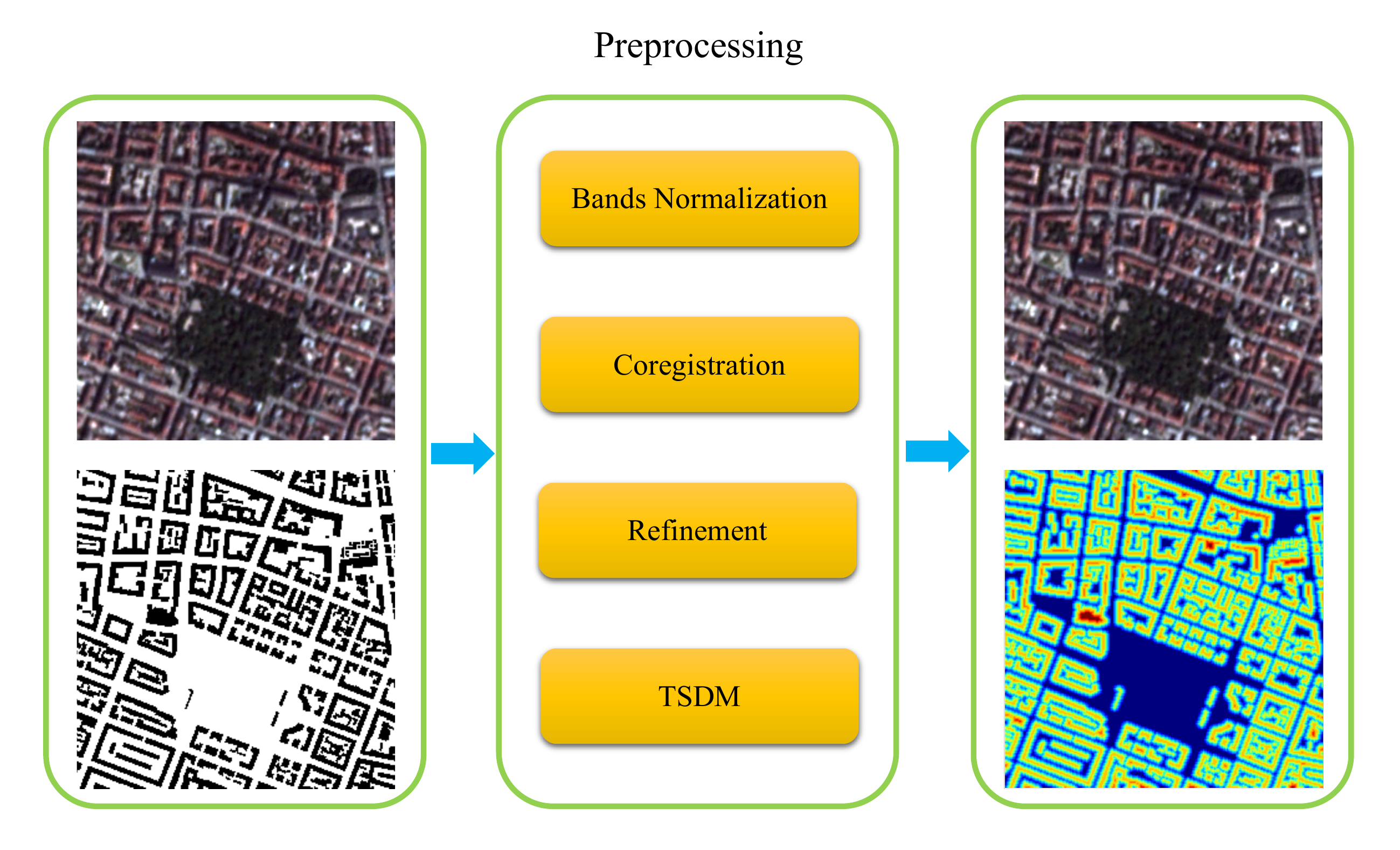}
  \caption{Illustration of preprocessing step}
  \label{fig:pre_processing}
\end{figure}

The datasets utilized in this work consist of Planetscope satellite imagery and OSM building footprints as ground truth. However, since data sources for OSM are different from satellite imagery, there are likely inconsistencies between OSM building footprints and satellite imagery. Therefore, we need to carry out preprocessing steps to limit the inconsistencies before the experiments, which include band normalization, coregistration, refinement, and a truncated signed distance map (TSDM) (see Fig. \ref{fig:pre_processing}).

In the next section, we will mainly focus on the coregistration and TSDM steps.

\subsubsection{Coregistration}
\begin{figure}[!ht]
  \centering
  \subfloat[]{\includegraphics[width=0.4\textwidth]{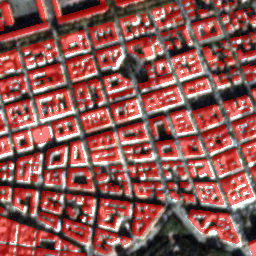}}
  \hfil
  \subfloat[]{\includegraphics[width=0.4\textwidth]{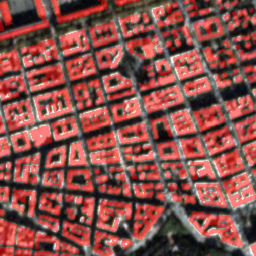}}
  \caption{ (a) Before coregistration; (b) After coregistration.}
  \label{fig:coregistration}
\end{figure}

One inconsistency is misalignments between OSM building footprints and satellite imagery, which is caused by different projections and accuracy levels from data sources. Fig. \ref{fig:coregistration} (a) shows an example of and OSM building footprint overlaid with the corresponding satellite imagery. There are noticeable misalignments between the building footprint and the satellite imagery. These misalignments lead to inaccurate training samples, which need to be corrected.

The coregistration process includes several steps: (1) The satellite imagery is transformed from RGB to gray scale; (2) The Gaussian gradient of grayscale imagery is calculated; (3) The cross correlation between the gradient magnitude of the grayscale image and building footprints is computed; (4) The pixel with the maximum cross correlation is found and the offset in both row and column direction can be derived. Fig. \ref{fig:coregistration} (b) shows the result after coregistration.

\subsubsection{Truncated signed distance map}
In order to incorporate both semantic information about class labels and geometric properties in the training of the network, the distances of pixels to boundaries of buildings are extracted as output representations. In our experiment, the value of the signed-distance function (SDF) is determined by the distance between the pixel and its nearest point on the boundary. Positive values imply that the pixels are within the buildings and negative values indicate the outside of buildings.

Then we truncate the distance at a given threshold to incorporate only the pixels closest to the border. In this case, the problem in our research is a multi-label segmentation task, which enhances the result of prediction by the detailed signed distance map. The truncated signed distance function can be expressed as:
\begin{equation}
  D(\mathbf{x}) = \delta_{d} \cdot \min \left( \min \limits_{\mathbf{x} \in \mathbf{X}} (d(\mathbf{x})), T_{d} \right),
\end{equation}
where $\min \limits_{\mathbf{x} \in \mathbf{X}} (d(\mathbf{x}))$ denotes the euclidean distance $d(\mathbf{x})$ between the pixel and its nearest point on the boundary of the building. The term $\delta_{d}$ is a sign function with the implication of inside or outside of objects; $T_{d}$ is the truncated threshold.

\subsection{Experimental setup}
We use 11 classes for the truncated signed distance map, which is in $[0, 10]$ and the truncated threshold is set to 5. For all networks, a stochastic gradient descent (SGD) is used and the learning rate is set to $\mathrm{10}^{-4}$. The negative log likelihood loss (NLLLoss) is adopted as the loss function. The proposed framework is implemented using Pytorch. Experiments are run on a NVIDIA Tesla P100 16 GB GPU. Several semantic segmentation methods, which include FCN-32s, SegNet, FCN-16s, U-Net, FCN-8s, ResNet-DUC, CWGAN-GP, FC-DenseNet, GCN, GraphSAGE, and GGNN, are chosen as the algorithms of comparison.

\subsection{Numerical results}
The three metrics in the following experiments selected to evaluate the results are: overall accuracy (OA), F1 scores, and the Intersection over Union (IoU) scores. The experiments are carried out in following way. First, as a baseline, we assess the capability of different deep convolutional neural networks for building footprint extraction. Then, we choose different DCNNs for deep structured feature embedding and combine it with GCN \cite{bib:kipf2016} to decide which DCNN is the best feature extractor for our proposed framework. At the end, we use the best feature extractor for DSFE and compare the proposed framework to different graph models. 

\subsubsection{Baseline with different DCNNs}
In this section, the performance of the state-of-the-art DCNNs for building footprint generation are firstly investigated, which indicates the capability of each DCNN for feature extraction and precise localization.

\begin{figure}
    \centering
    \begin{tabular}{cccc}
    \subfloat[]{\includegraphics[width=0.24\textwidth]{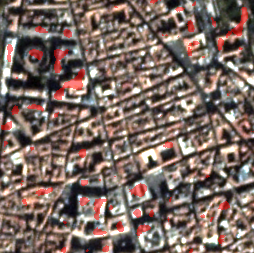}} &
    \subfloat[]{\includegraphics[width=0.24\textwidth]{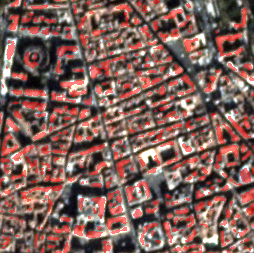}} &
    \subfloat[]{\includegraphics[width=0.24\textwidth]{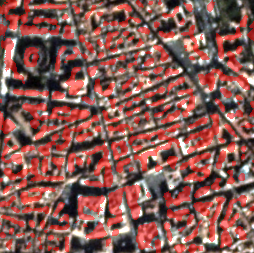}} &
    \subfloat[]{\includegraphics[width=0.24\textwidth]{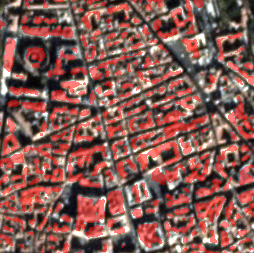}} \\

    \subfloat[]{\includegraphics[width=0.24\textwidth]{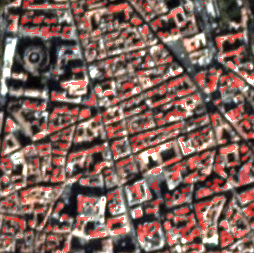}} &
    \subfloat[]{\includegraphics[width=0.24\textwidth]{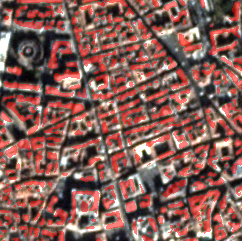}} &
    \subfloat[]{\includegraphics[width=0.24\textwidth]{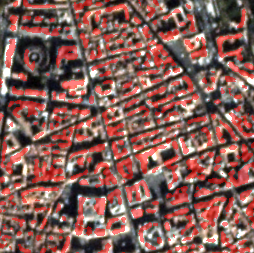}} &
    \subfloat[]{\includegraphics[width=0.24\textwidth]{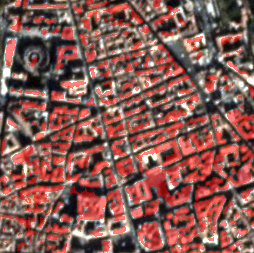}} \\

    \subfloat[]{\includegraphics[width=0.24\textwidth]{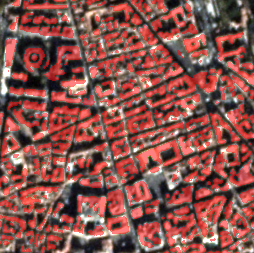}} &
    \subfloat[]{\includegraphics[width=0.24\textwidth]{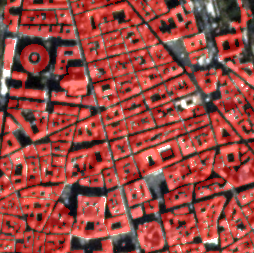}} &  & \\
    \end{tabular}
    \caption{Visualized comparison of the predicted results using different DCNNs. The predicted label is red, which overlays the optical image. (a) FCN-32s; (b) FCN-16s; (c) ResNet-DUC; (d) E-Net; (e) SegNet; (f) U-Net; (g) FCN-8s; (h) CWGAN-GP; (i) FC-DenseNet; (j) Ground truth. }
    \label{fig:comp_different_dcnns_methods}
\end{figure}

\begin{table}[ht]
\begin{center}
\begin{tabular}{cccc}
\toprule
\toprule
Methods  & \textbf{OA} & \textbf{F1} & \textbf{IoU} \\
\midrule
\rule{0pt}{2.5ex} FCN-32s  & 0.7318 & 0.2697 & 0.1559  \\
\rule{0pt}{2.5ex} FCN-16s  & 0.7698 & 0.3993 & 0.2494  \\
\rule{0pt}{2.5ex} ResNet-DUC  & 0.7945 & 0.4542 & 0.2930  \\
\rule{0pt}{2.5ex} E-Net  & 0.8243  & 0.5427 & 0.3724 \\
\rule{0pt}{2.5ex} SegNet & 0.8261  & 0.5558 & 0.3848 \\
\rule{0pt}{2.5ex} U-Net & 0.8412 & 0.6043  & 0.4329 \\
\rule{0pt}{2.5ex} FCN-8s & 0.8472 & 0.6222 & 0.4513 \\
\rule{0pt}{2.5ex} CWGAN-GP  & 0.8483 & 0.6268 & 0.4562 \\
\rule{0pt}{2.5ex} FC-DenseNet & \textbf{0.8551} & \textbf{0.6328} & \textbf{0.4628}  \\
\bottomrule
\bottomrule
\end{tabular}
\end{center}
\caption{Comparison of different deep convolutional neural networks on the test datasets}
\label{tab:comp_dcnn_results}
\end{table}

FCN-32s and FCN-16s exhibit poor performance, since the feature map of later layers have only high-level semantics with poor localization. ResNet-DUC can achieve better result than the previous two because of hybrid dilated convolution and dense upsampling convolution. However, it is limited due to the lack of skip connections. Max-pooling indices are reused in SegNet during the decoding process, which can reduce the parameter number of network leading to efficient training. However, as it only use indices of max-pooling to decoder, some local details cannot be recovered, e.g., small buildings will be neglected. FCN-8s and U-Net outperform previous networks due to the concatenation of low-level features. Compared to the other CNN models, cwGAN-gp shows promising results for building footprint generation. The enhancement of performance is motivated by the min-max competition between the discriminator and the generator of the GAN.

FC-DenseNet outperforms all other semantic segmentation neural networks in numerical accuracy and visual results. On one hand, DenseNet block concatenates different features learned by convolution layers, which can boost the input diversity of subsequent layers and promote better efficiency of the training. On the other hand, the detailed spatial information can be propagated by shortcut connections between the convolution and deconvolution paths, which enhances the recovery of fine-grained segmentation from the deconvolution path.

\subsubsection{Proposed framework with different DSFE}
In order to choose the best feature extractor for our task, three representative DCNNs have been adopted in the proposed framework with the graph convolutional network. The statistical result is shown in Table \ref{tab:comp_diff_dsfe_gcn_results}.

\begin{table}[ht]
\begin{center}
\begin{tabular}{cccc}
\toprule
\toprule
Methods  & \textbf{OA} & \textbf{F1} & \textbf{IoU} \\
\midrule
\rule{0pt}{2.5ex} DSFE(U-Net)-GCN & 0.8396 & 0.6258  & 0.4544  \\
\rule{0pt}{2.5ex} DSFE(FCN-8s)-GCN & 0.8594 & 0.6320 & 0.4611  \\
\rule{0pt}{2.5ex} DSFE(FC-DenseNet)-GCN & \textbf{0.8640} & \textbf{0.6677} & \textbf{0.5012}  \\
\bottomrule
\bottomrule
\end{tabular}
\end{center}
\caption{Quantitative comparison of different deep neural networks on Planetscope's datasets}
\label{tab:comp_diff_dsfe_gcn_results}
\end{table}

From Table \ref{tab:comp_diff_dsfe_gcn_results} we can see that different DCNNs exhibit different capabilities for feature embedding. It is clear that FC-DenseNet, as a feature extractor in DSFE with GCN, produces the best result. This is due to the superiority of FC-DenseNet, which extends the DenseNet architecture to a U-Net-like network for semantic segmentation. In the DenseNet block, through feature reuse, there are shorter connections between layers close to the input and those close to the output, which force the intermediate layers to learn discriminative features. Moreover, DenseNet combines features by iteratively concatenating them, which contributes to improved information and gradient propagation in the networks.

\begin{figure}[!ht]
  \centering
  \subfloat[]{\includegraphics[width=0.4\textwidth]{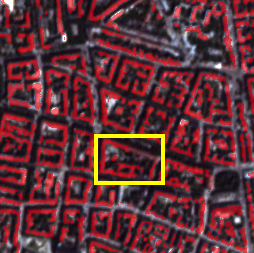}}
  \hfil
  \subfloat[]{\includegraphics[width=0.4\textwidth]{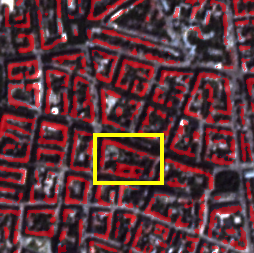}}
  \vfil
  \subfloat[]{\includegraphics[width=0.4\textwidth]{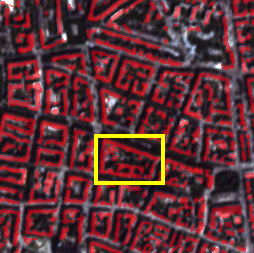}}
  \hfil
  \subfloat[]{\includegraphics[width=0.4\textwidth]{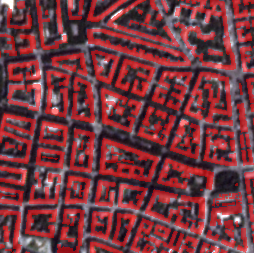}}
  \caption{Visualized comparison of the predicted results using different DCNNs in DSFE with GCN. The yellow bounding box highlights the key region for comparison. (a) DSFE (U-Net)-GCN; (b) DSFE (FCN-8s)-GCN; (c) DSFE (FC-DenseNet)-GCN; (d) Ground truth. }
  \label{fig:comp_different_dsfe}
\end{figure}
As can be seen in Fig. \ref{fig:comp_different_dsfe}, DSFE (FC-DenseNet)-GCN gives the best result, which implies that FC-DenseNet is a powerful tool for extracting different levels of features.

\subsubsection{Proposed framework with different graph models}

In this section, we choose FC-DenseNet as the feature extractor in DSFE with different graph models. The results are summarized in Table \ref{tab:comp_dsfe_gcn_results}.
\begin{table}[ht]
\begin{center}
\begin{tabular}{cccc}
\toprule
\toprule
Methods  & \textbf{OA} & \textbf{F1} & \textbf{IoU} \\
\midrule
\rule{0pt}{2.5ex} FC-DenseNet \cite{bib:jegou2017} & 0.8551 & 0.6328  & 0.4628  \\
\rule{0pt}{2.5ex} DSFE-CRF \cite{bib:chen2017} & 0.8592 & 0.6415  & 0.4757  \\
\rule{0pt}{2.5ex} DSFE-GCN \cite{bib:kipf2016} & 0.8640 & 0.6677 & 0.5012  \\
\rule{0pt}{2.5ex} DSFE-GraphSAGE \cite{bib:hamilton2017} & 0.8719 & 0.6726 & 0.5067  \\
\rule{0pt}{2.5ex} DSFE-GGNN \cite{bib:li2016} & 0.8787 & 0.6778 & 0.5123  \\
\rule{0pt}{2.5ex} DSFE-GGCN & \textbf{0.8881} & \textbf{0.6899} & \textbf{0.5251} \\
\bottomrule
\bottomrule
\end{tabular}
\end{center}
\caption{Comparison of different networks on the Planetscope dataset}
\label{tab:comp_dsfe_gcn_results}
\end{table}

\begin{figure}
    \centering
    \begin{tabular}{cccc}
    \subfloat[]{\includegraphics[width=0.24\textwidth]{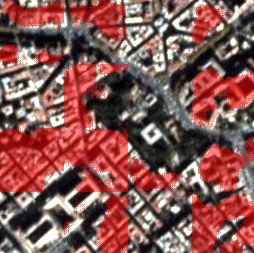}} &
    \subfloat[]{\includegraphics[width=0.24\textwidth]{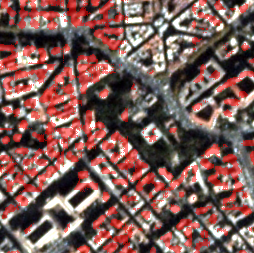}} &
    \subfloat[]{\includegraphics[width=0.24\textwidth]{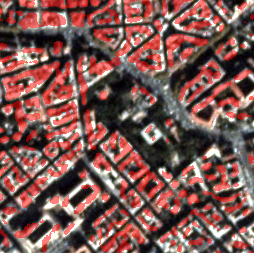}} &
    \subfloat[]{\includegraphics[width=0.24\textwidth]{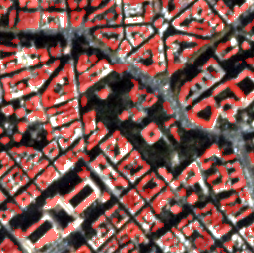}} \\

    \subfloat[]{\includegraphics[width=0.24\textwidth]{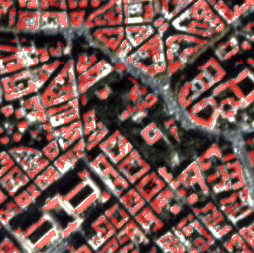}} &
    \subfloat[]{\includegraphics[width=0.24\textwidth]{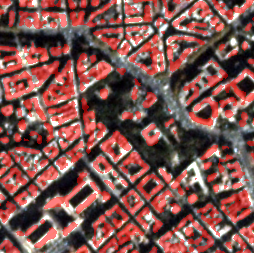}} &
    \subfloat[]{\includegraphics[width=0.24\textwidth]{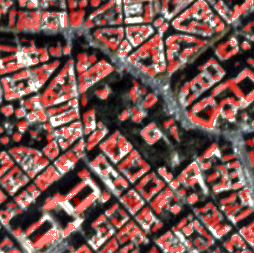}} &
    \subfloat[]{\includegraphics[width=0.24\textwidth]{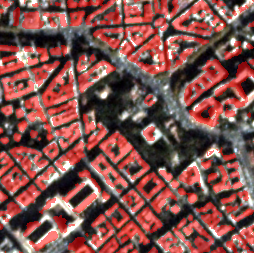}} \\

    \subfloat[]{\includegraphics[width=0.24\textwidth]{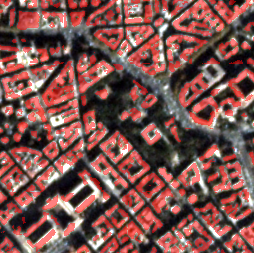}} &
    \subfloat[]{\includegraphics[width=0.24\textwidth]{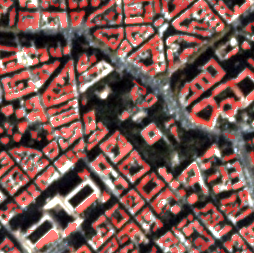}} &
    \subfloat[]{\includegraphics[width=0.24\textwidth]{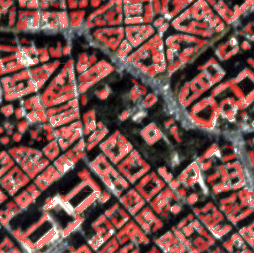}} &
    \subfloat[]{\includegraphics[width=0.24\textwidth]{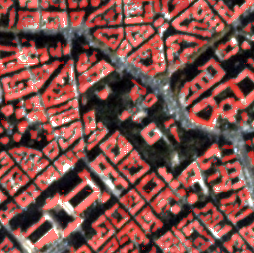}} \\

    \subfloat[]{\includegraphics[width=0.24\textwidth]{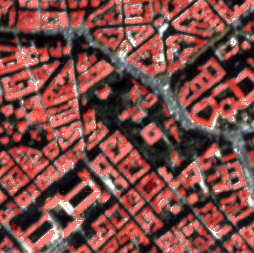}} &
    \subfloat[]{\includegraphics[width=0.24\textwidth]{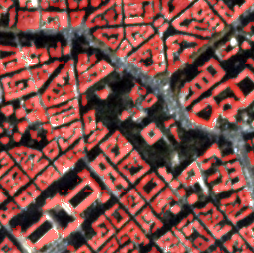}} &
    \subfloat[]{\includegraphics[width=0.24\textwidth]{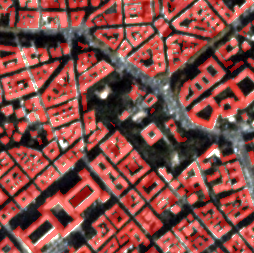}} &
    \subfloat[]{\includegraphics[width=0.24\textwidth]{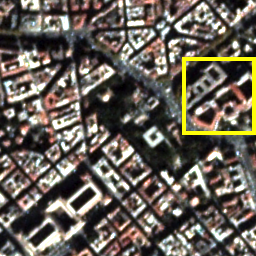}} \\
    \end{tabular}
    \caption{Visualized comparison of the predicted results using different networks. (a) FCN-32; (b) FCN-16s; (c) ResNet-DUC; (d) E-Net; (e) SegNet; (f) U-Net; (g) FCN-8s; (h) CWGAN-GP; (i) FC-DenseNet; (j) DSFE-CRF; (k) DSFE-GCN; (l) DSFE-GraphSAGE; (m) DSFE-GGNN; (n) DSFE-GGCN; (o) Ground truth; (p) Optical image.}
    \label{fig:comp_different_methods_planet_large}
\end{figure}

\begin{figure}
    \centering
    \begin{tabular}{cccc}
    \subfloat[]{\includegraphics[width=0.24\textwidth]{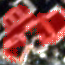}} &
    \subfloat[]{\includegraphics[width=0.24\textwidth]{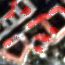}} &
    \subfloat[]{\includegraphics[width=0.24\textwidth]{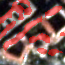}} &
    \subfloat[]{\includegraphics[width=0.24\textwidth]{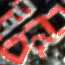}} \\

    \subfloat[]{\includegraphics[width=0.24\textwidth]{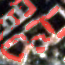}} &
    \subfloat[]{\includegraphics[width=0.24\textwidth]{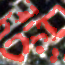}} &
    \subfloat[]{\includegraphics[width=0.24\textwidth]{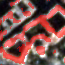}} &
    \subfloat[]{\includegraphics[width=0.24\textwidth]{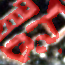}} \\

    \subfloat[]{\includegraphics[width=0.24\textwidth]{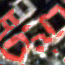}} &
    \subfloat[]{\includegraphics[width=0.24\textwidth]{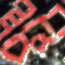}} &
    \subfloat[]{\includegraphics[width=0.24\textwidth]{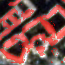}} &
    \subfloat[]{\includegraphics[width=0.24\textwidth]{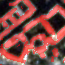}} \\

    \subfloat[]{\includegraphics[width=0.24\textwidth]{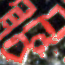}} &
    \subfloat[]{\includegraphics[width=0.24\textwidth]{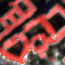}} &
    \subfloat[]{\includegraphics[width=0.24\textwidth]{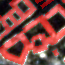}} &
    \subfloat[]{\includegraphics[width=0.24\textwidth]{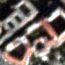}} \\
    \end{tabular}
    \caption{Visualized comparison of the predicted results using different networks. (a) FCN-32; (b) FCN-16s; (c) ResNet-DUC; (d) E-Net; (e) SegNet; (f) U-Net; (g) FCN-8s; (h) CWGAN-GP; (i) FC-DenseNet; (j) DSFE-CRF; (k) DSFE-GCN; (l) DSFE-GraphSAGE; (m) DSFE-GGNN; (n) DSFE-GGCN; (o) Ground truth; (p) Optical image.}
    \label{fig:comp_different_methods_planet_small}
\end{figure}

The results show that DSFE-GGCN has the best performance for our task. The IoU increases 6.2\% compared to the best result of DCNN. Fig. \ref{fig:comp_different_methods_planet_large} shows a visual comparison of all the networks used in section \ref{section4}. We marked the key region with a yellow bounding box. The close-up figures for the key regions are shown in Fig. \ref{fig:comp_different_methods_planet_small}.

\section{Discussion}
\subsection{Additional dataset}

We validate our proposed method with experiments on the ISPRS 2D Semantic Labeling Contest dataset, which covers the city of Potsdam and comprises 38 tiles of aerial imagery \cite{bib:isprs}. In order to maintain the consistency, images with 3 spectral bands (red, green, blue) are used in this experiment without a digital surface model (DSM). Each aerial image is depicted with $6000 \times 6000$ pixels at a spatial resolution of 5 cm. The corresponding ground truth is also provided for results evaluation, which includes six classes: Impervious surfaces, Building, Low vegetation, Trees, Cars, and Clutter/background. For our detailed experiments, we split those 38 tiles into a training subset (tile numbers 2-10 to 6-15) and a test subset (tile numbers 7-07 to 7-13). The building class is regarded as a building and other five classes are considered non-buildings. We cut 16,000 patches of $256 \times 256$ pixels from the training subset and 3573 patches from the test subset. As mentioned in the previous section, the data augmentation step TSDM is used for the medium-resolution images and the ground truth is well coregistrated with the optical image. Therefore, there is no data preprocessing step for the ISPRS dataset. The optical image is fed directly into the networks.

\subsection{Experimental setup}
The SGD optimizer is adopted and the initial learning rate is set to be 10e-4, which is reduced by a factor of ten when the validation loss is saturated. Once the learning rate is reduced below 10e-8, the training stops. The number of epochs is in the range (120, 160) for all the networks. The size of the training batch is 4.

\subsection{Experimental results}
The metrics OA, F1 scores, and IoU scores are used to evaluate the results. Fig. \ref{fig:comp_different_methods_isprs_large} shows the visualized comparison of the predicted results the ISPRS Potsdam dataset, using different networks.

FCN-8s provides a significantly higher percentage of buildings detected compared to FCN-16s and FCN-32s, by combining predictions from not only the final layer but also coarse layers, allowing more information to be preserved. The boundaries of buildings detected from U-Net are sharper than for SegNet or E-Net. However, unlike in the medium-resolution case, the completeness of the result obtained by SegNet or E-Net is better than for U-Net, which indicates that the spatial information propagation is more effectively undertaken by recording the pooling indices than by concatenating the low-level features when the resolution is high enough, i.e., when comprehensive spatial information exists.
The finer details are captured by the proposed framework with different graph models such as CRFasRNN, GCN, and GGCN rather than CNN-only methods, which confirms the effectiveness of the graph model in modelling the interaction among pixels and spatial information propagation. Compared to CRFasRNN and GCN, the proposed GGCN method gives a better result. 
A close-up view of the key region is shown in Fig. \ref{fig:comp_different_methods_isprs_small}. It can be seen that the DSFE-GGCN shows a better result with respect to both completeness and sharpness for building extraction compared to other methods.

\begin{figure}
    \centering
    \begin{tabular}{cccc}
    \subfloat[]{\includegraphics[width=0.45\textwidth]{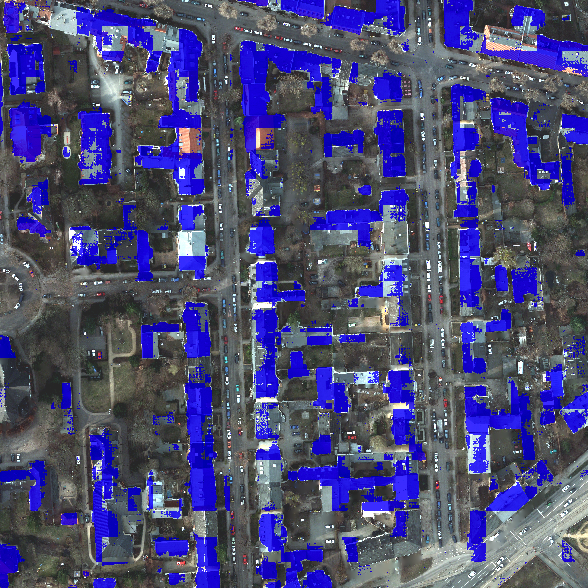}} &
    \subfloat[]{\includegraphics[width=0.45\textwidth]{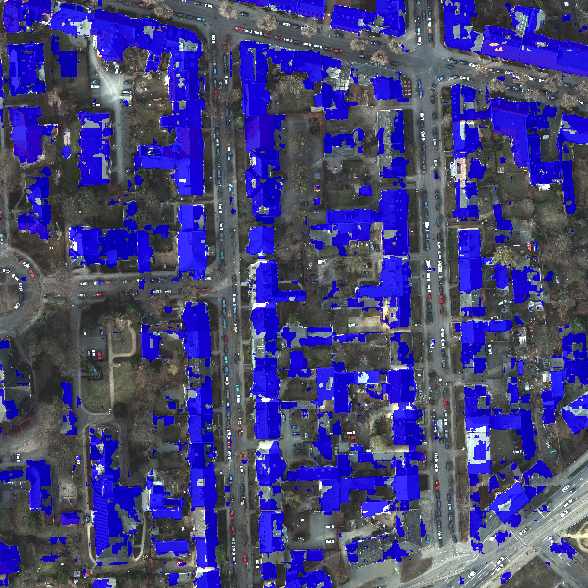}} \\

    \subfloat[]{\includegraphics[width=0.45\textwidth]{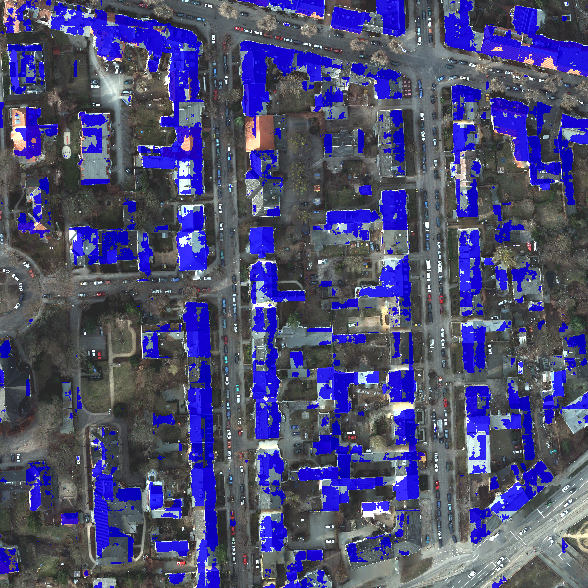}} &
    \subfloat[]{\includegraphics[width=0.45\textwidth]{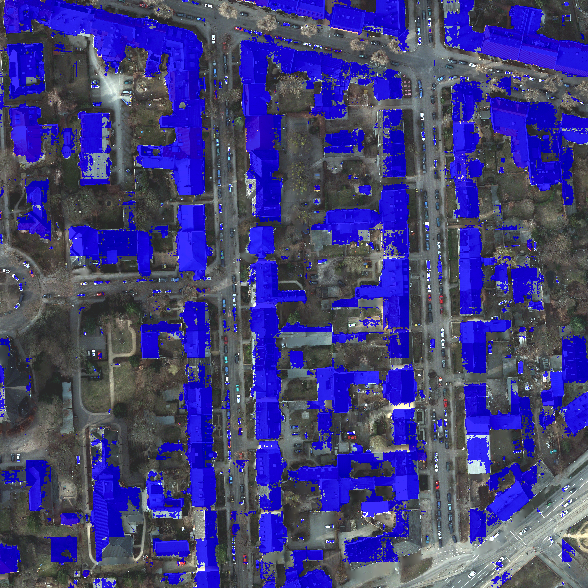}} \\

    \subfloat[]{\includegraphics[width=0.45\textwidth]{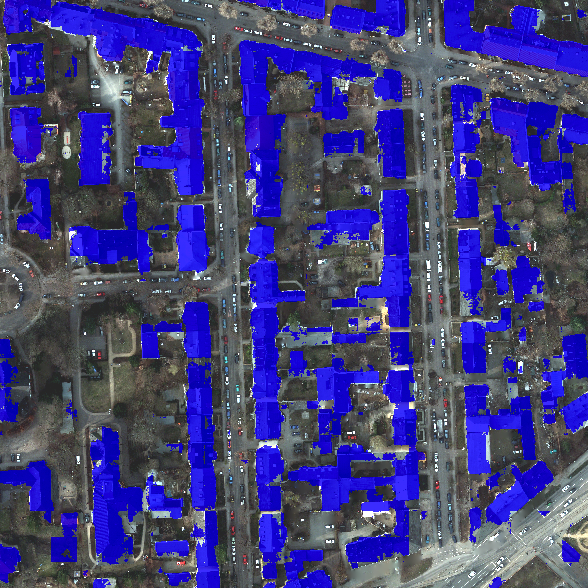}} &
    \subfloat[]{\includegraphics[width=0.45\textwidth]{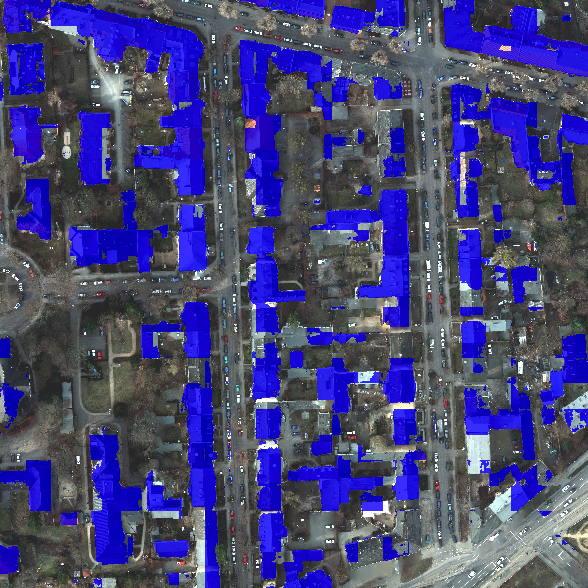}} \\

    \end{tabular}
    \caption{Visualized comparison of the predicted results from the ISPRS Potsdam dataset using different networks. (a) FCN-32; (b) FCN-16s; (c) ResNet-DUC; (d) E-Net; (e) SegNet; (f) U-Net; (g) FCN-8s; (h) CWGAN-GP; (i) FC-DenseNet; (j) DSFE-CRF; (k) DSFE-GCN; (l) DSFE-GGCN; (m) Ground truth; (n) Optical image.}
    \label{fig:comp_different_methods_isprs_large}
\end{figure}

\begin{figure}
  \ContinuedFloat
    \centering
    \begin{tabular}{cccc}
    \subfloat[]{\includegraphics[width=0.45\textwidth]{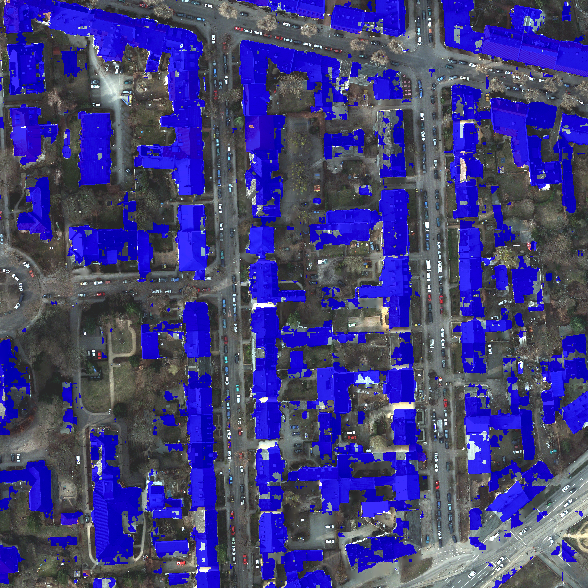}} &
    \subfloat[]{\includegraphics[width=0.45\textwidth]{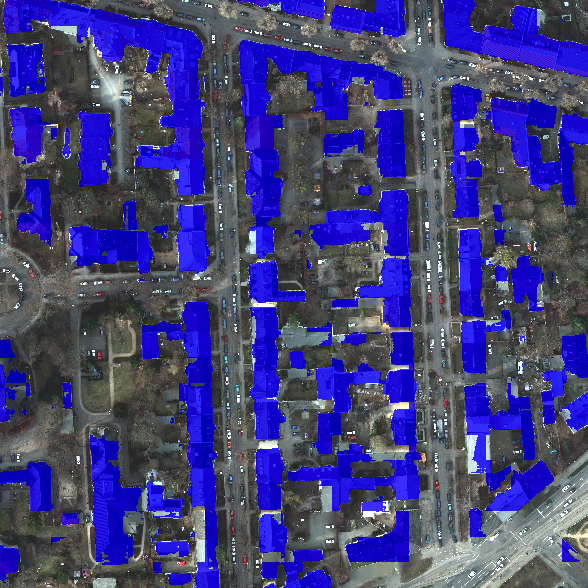}} \\

    \subfloat[]{\includegraphics[width=0.45\textwidth]{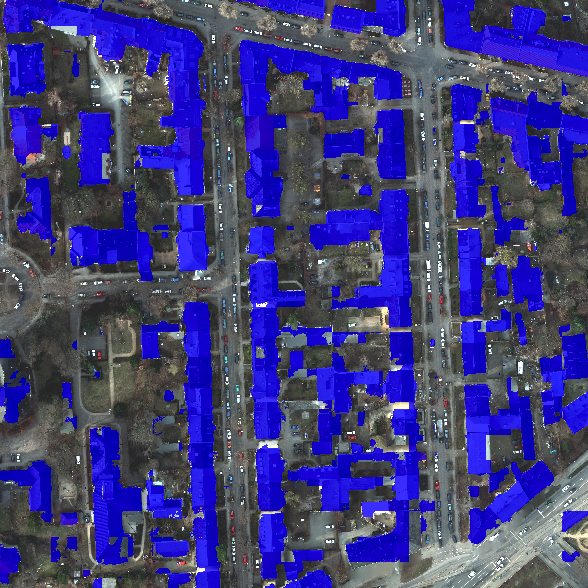}} &
    \subfloat[]{\includegraphics[width=0.45\textwidth]{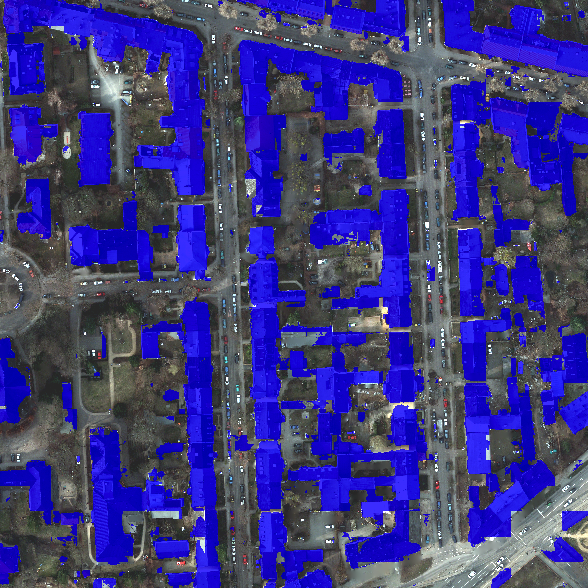}} \\

    \subfloat[]{\includegraphics[width=0.45\textwidth]{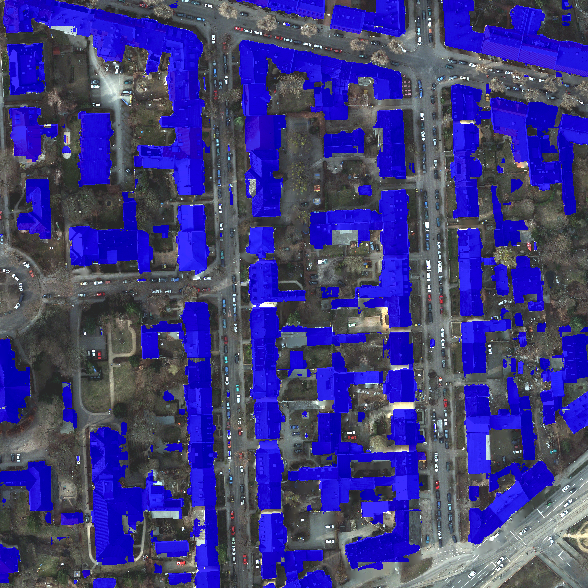}} &
    \subfloat[]{\includegraphics[width=0.45\textwidth]{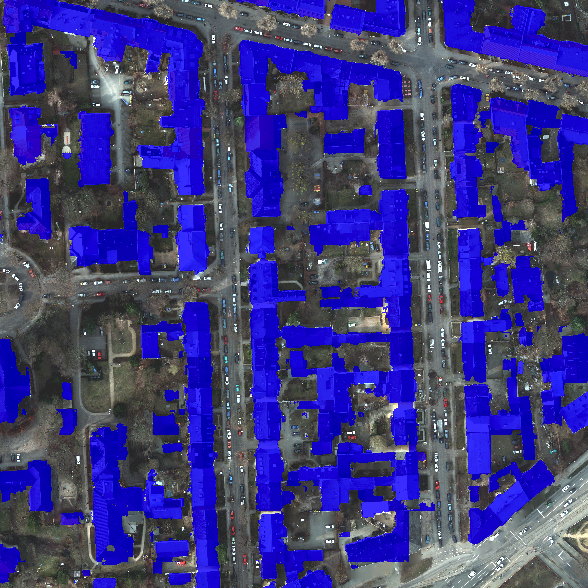}} \\

    \end{tabular}
    \caption{Visualized comparison of the predicted results from the ISPRS Potsdam dataset using different networks. (a) FCN-32; (b) FCN-16s; (c) ResNet-DUC; (d) E-Net; (e) SegNet; (f) U-Net; (g) FCN-8s; (h) CWGAN-GP; (i) FC-DenseNet; (j) DSFE-CRF; (k) DSFE-GCN; (l) DSFE-GGCN; (m) Ground truth; (n) Optical image.}
    \label{fig:comp_different_methods_isprs_large}
\end{figure}

\begin{figure}
  \ContinuedFloat
    \centering
    \begin{tabular}{cccc}
    \subfloat[]{\includegraphics[width=0.45\textwidth]{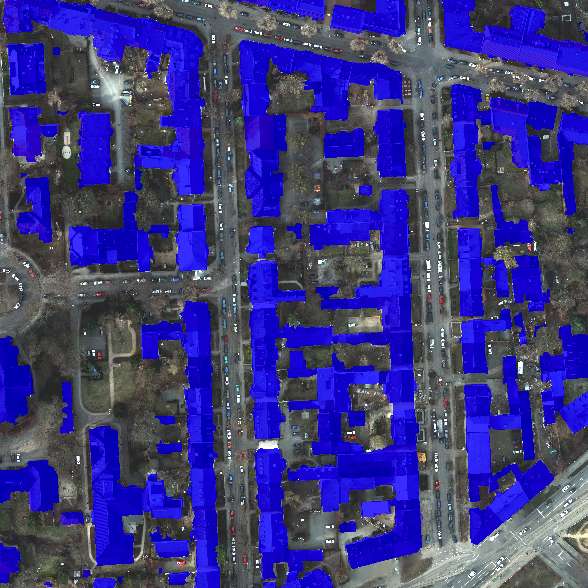}} &
    \subfloat[]{\includegraphics[width=0.45\textwidth]{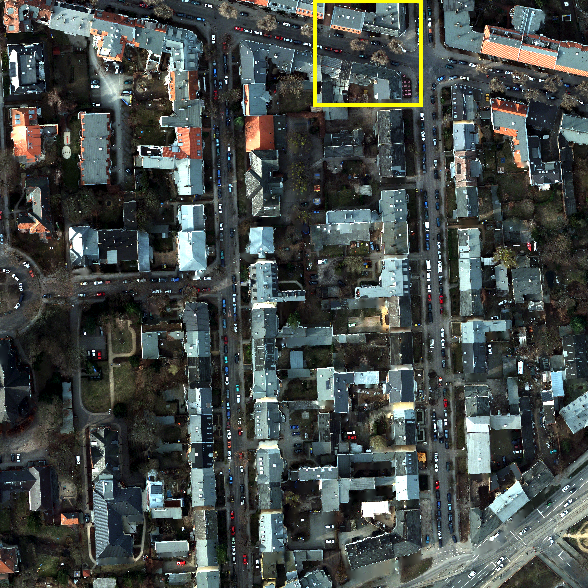}} & \\
    \end{tabular}
    \caption{Visualized comparison of the predicted results from the ISPRS Potsdam dataset using different networks. (a) FCN-32; (b) FCN-16s; (c) ResNet-DUC; (d) E-Net; (e) SegNet; (f) U-Net; (g) FCN-8s; (h) CWGAN-GP; (i) FC-DenseNet; (j) DSFE-CRF; (k) DSFE-GCN; (l) DSFE-GGCN; (m) Ground truth; (n) Optical image.}
    \label{fig:comp_different_methods_isprs_large}
\end{figure}

\begin{figure}
    \centering
    \begin{tabular}{cccc}
    \subfloat[]{\includegraphics[width=0.45\textwidth]{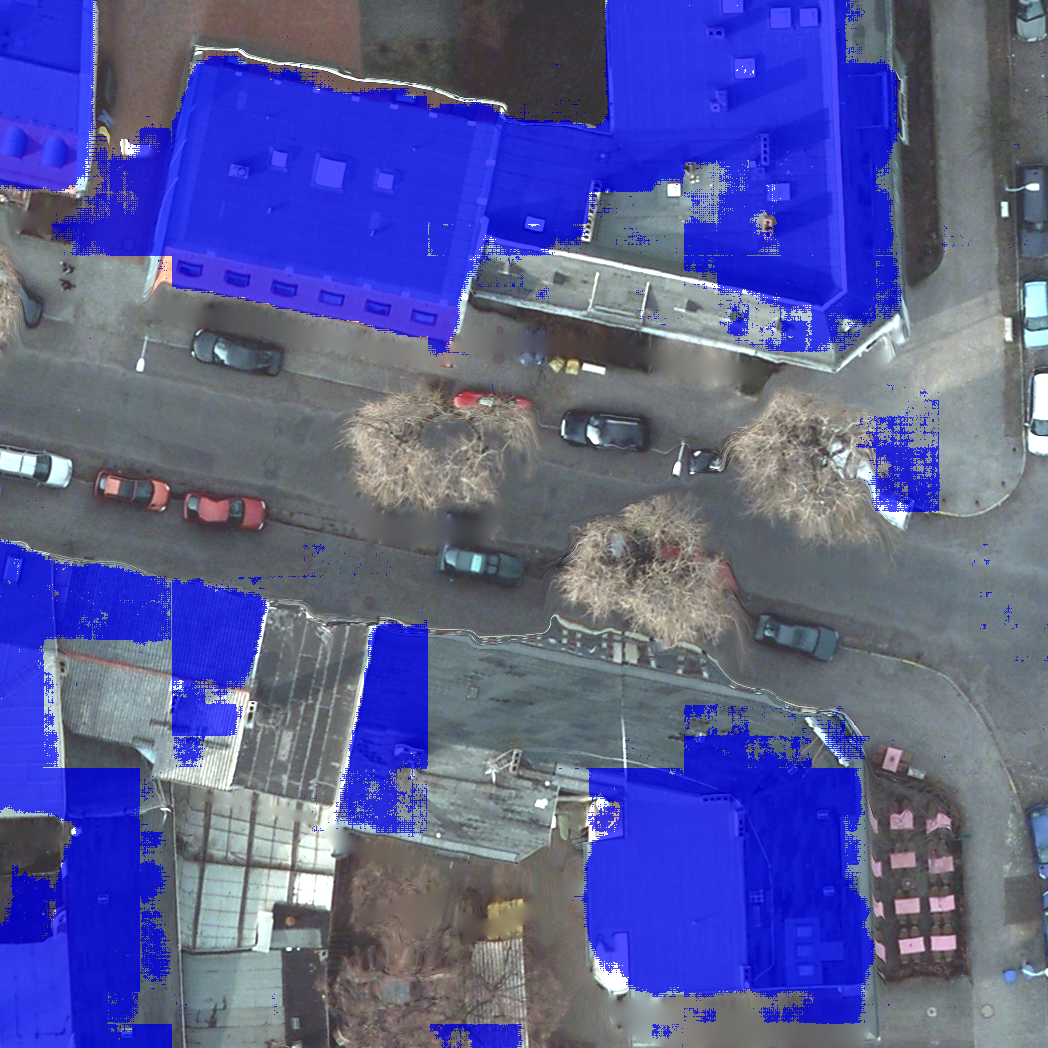}} &
    \subfloat[]{\includegraphics[width=0.45\textwidth]{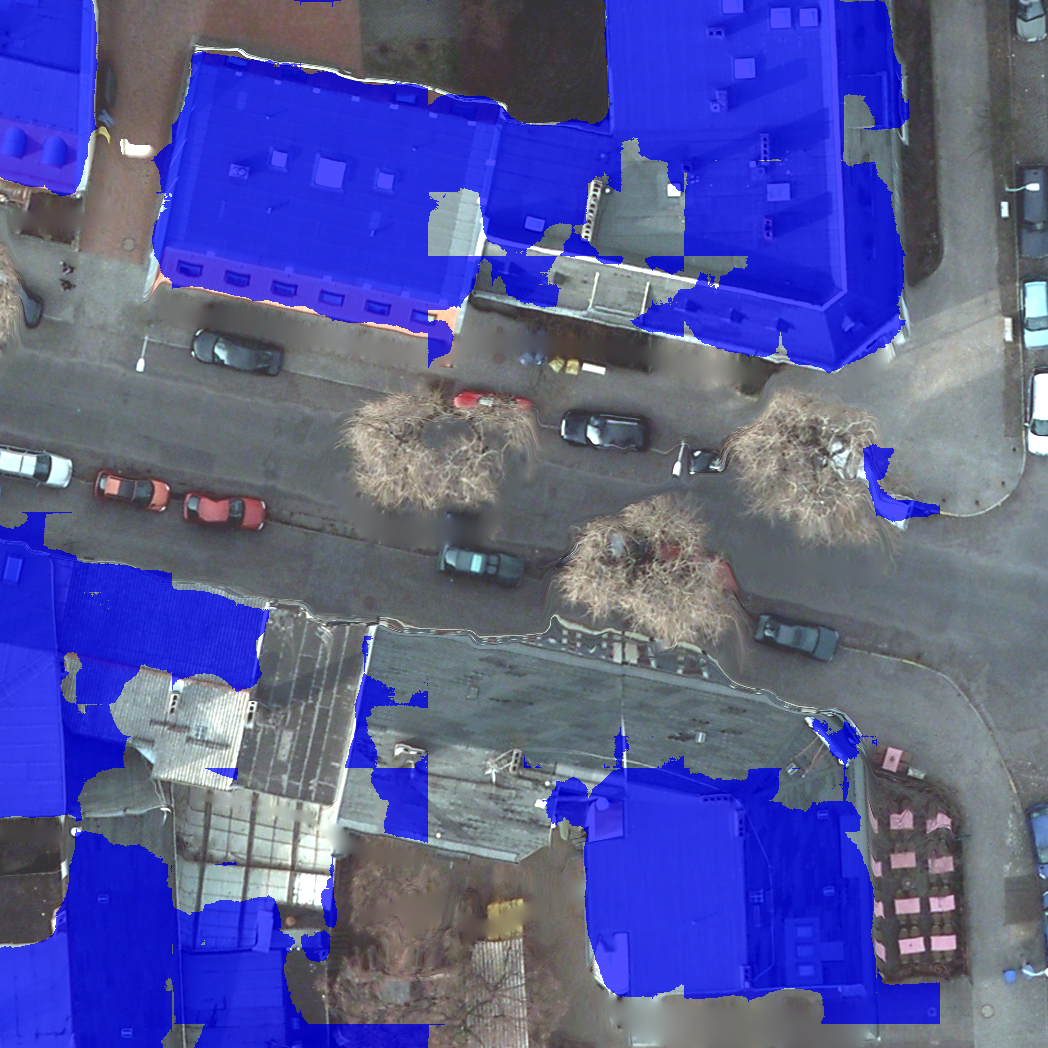}} \\

    \subfloat[]{\includegraphics[width=0.45\textwidth]{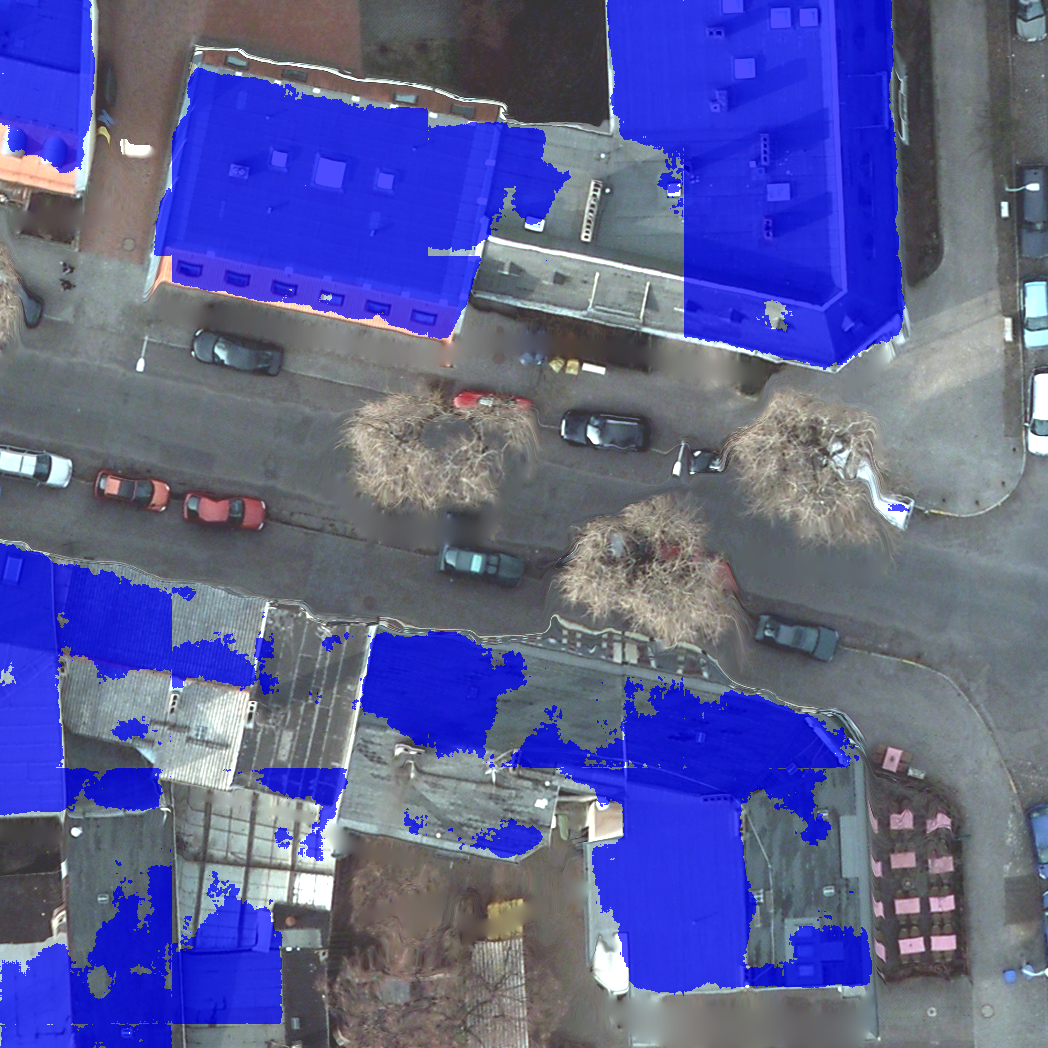}} &
    \subfloat[]{\includegraphics[width=0.45\textwidth]{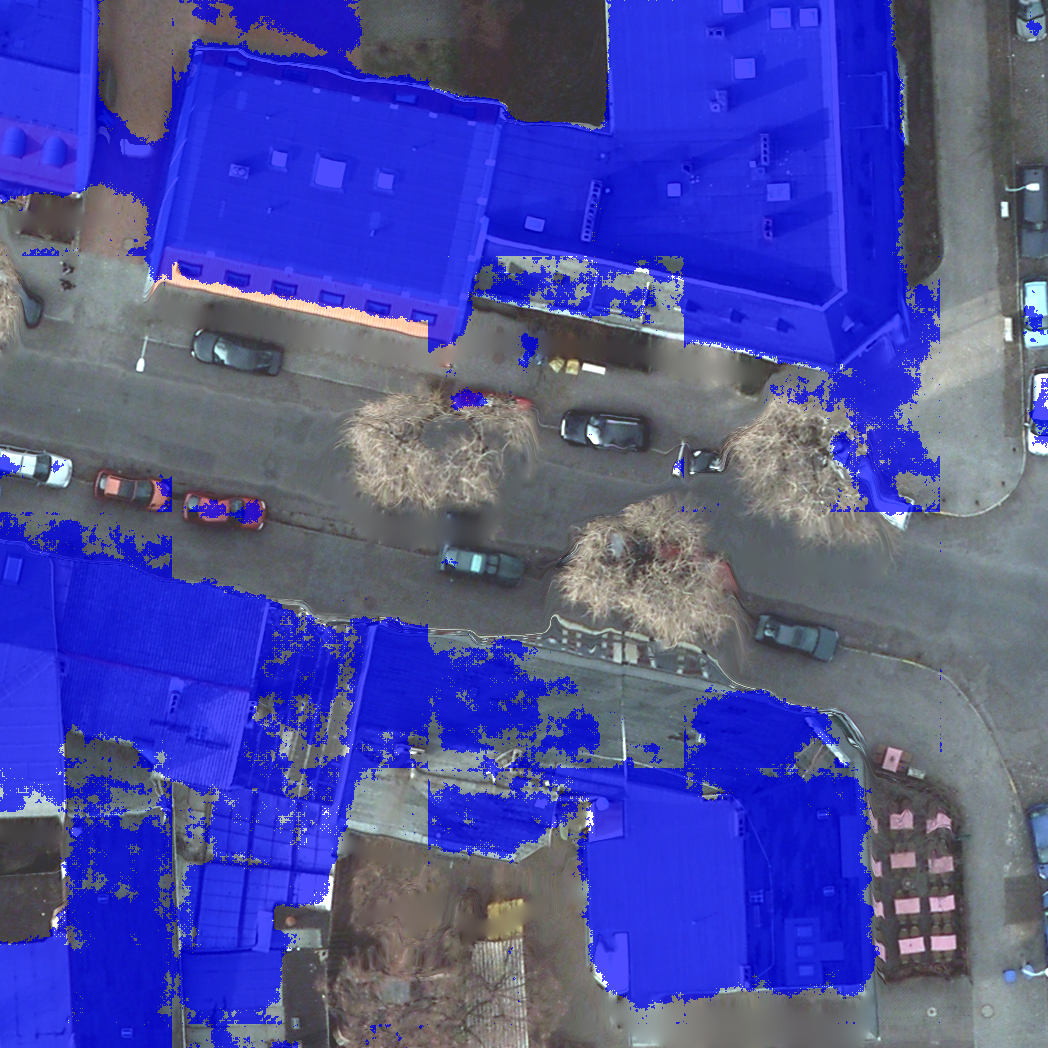}} \\

    \subfloat[]{\includegraphics[width=0.45\textwidth]{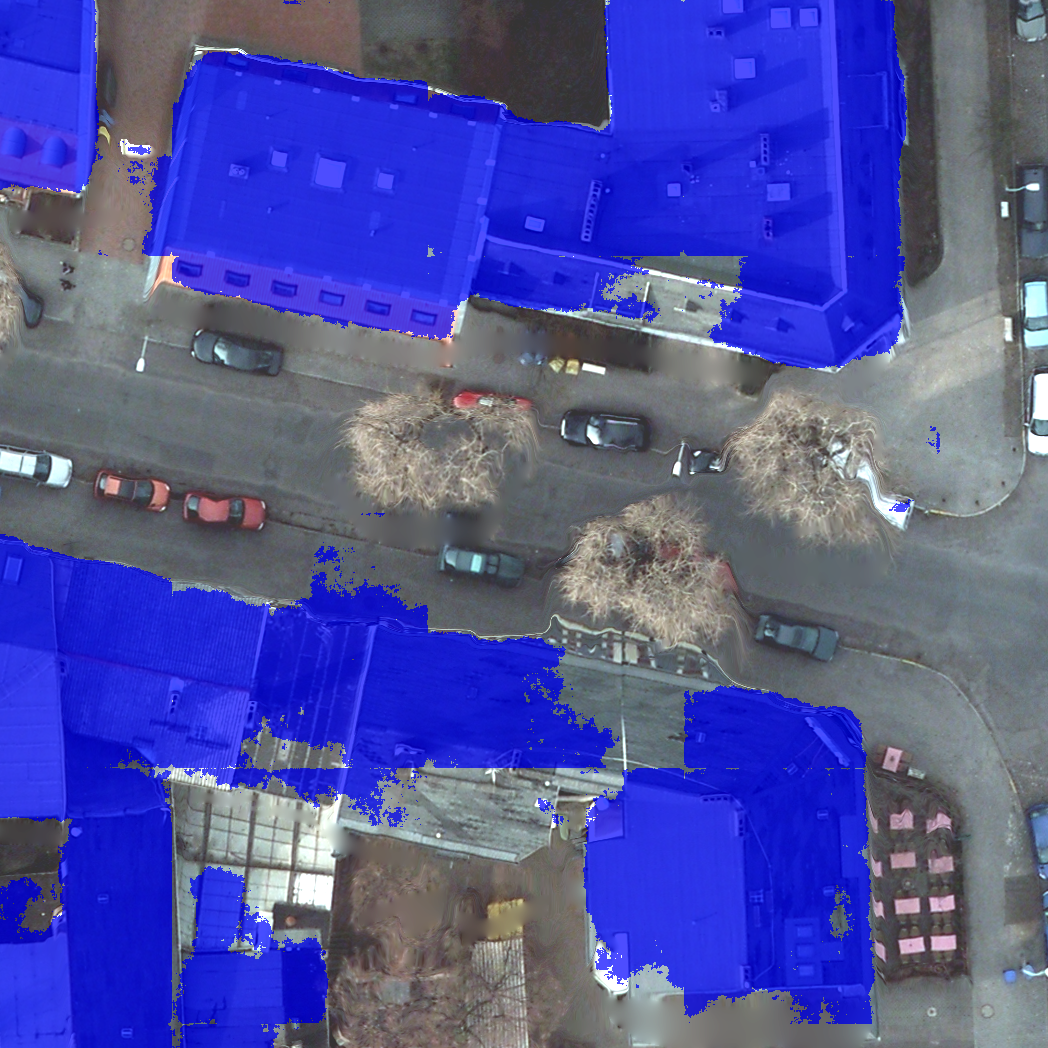}} &
    \subfloat[]{\includegraphics[width=0.45\textwidth]{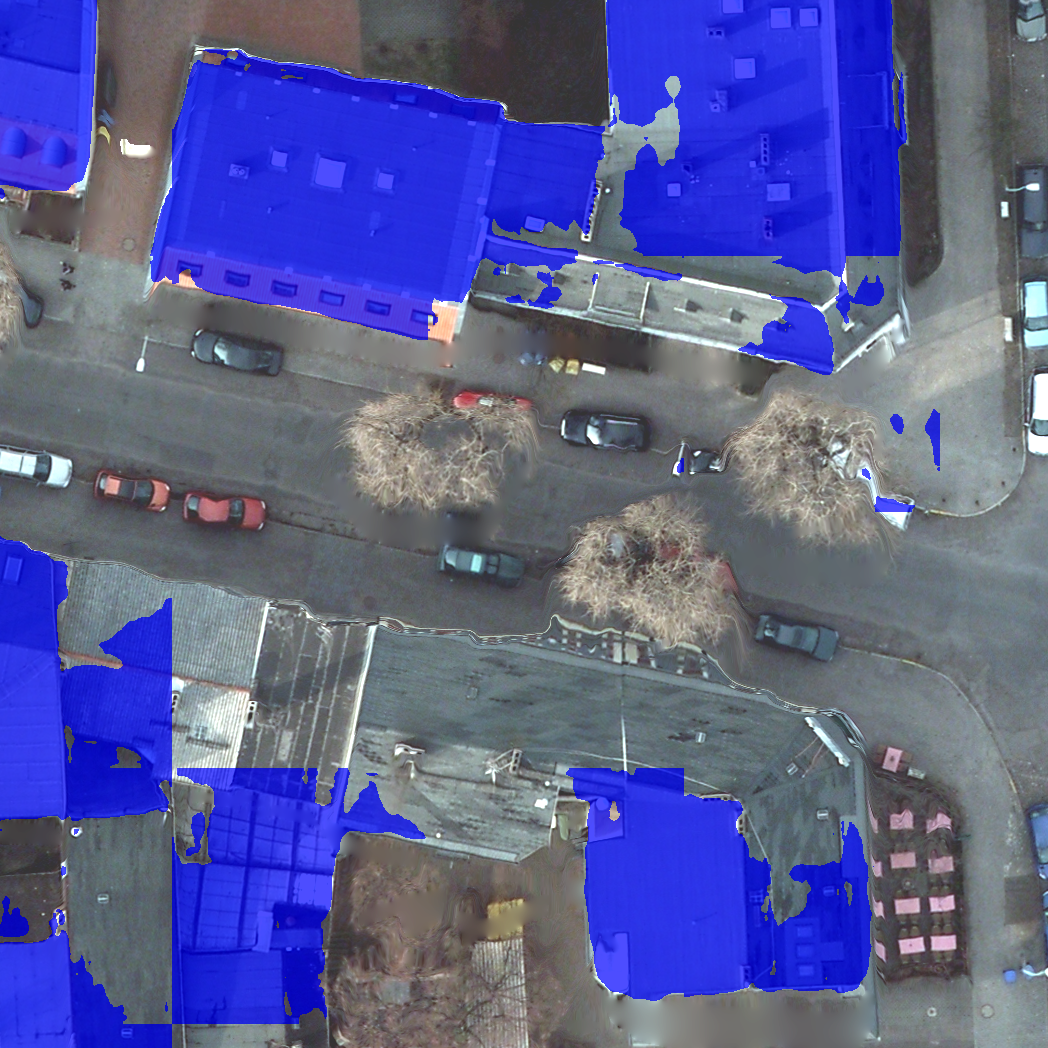}} \\

    \end{tabular}
    \caption{Visualized comparison of the predicted results on the ROI of ISPRS Potsdam dataset using different networks. (a) FCN-32; (b) FCN-16s; (c) ResNet-DUC; (d) E-Net; (e) SegNet; (f) U-Net; (g) FCN-8s; (h) CWGAN-GP; (i) FC-DenseNet; (j) DSFE-CRF; (k) DSFE-GCN; (l) DSFE-GGCN; (m) Ground truth; (n) Optical image.}
    \label{fig:comp_different_methods_isprs_small}
\end{figure}

\begin{figure}
  \ContinuedFloat
    \centering
    \begin{tabular}{cccc}
    \subfloat[]{\includegraphics[width=0.45\textwidth]{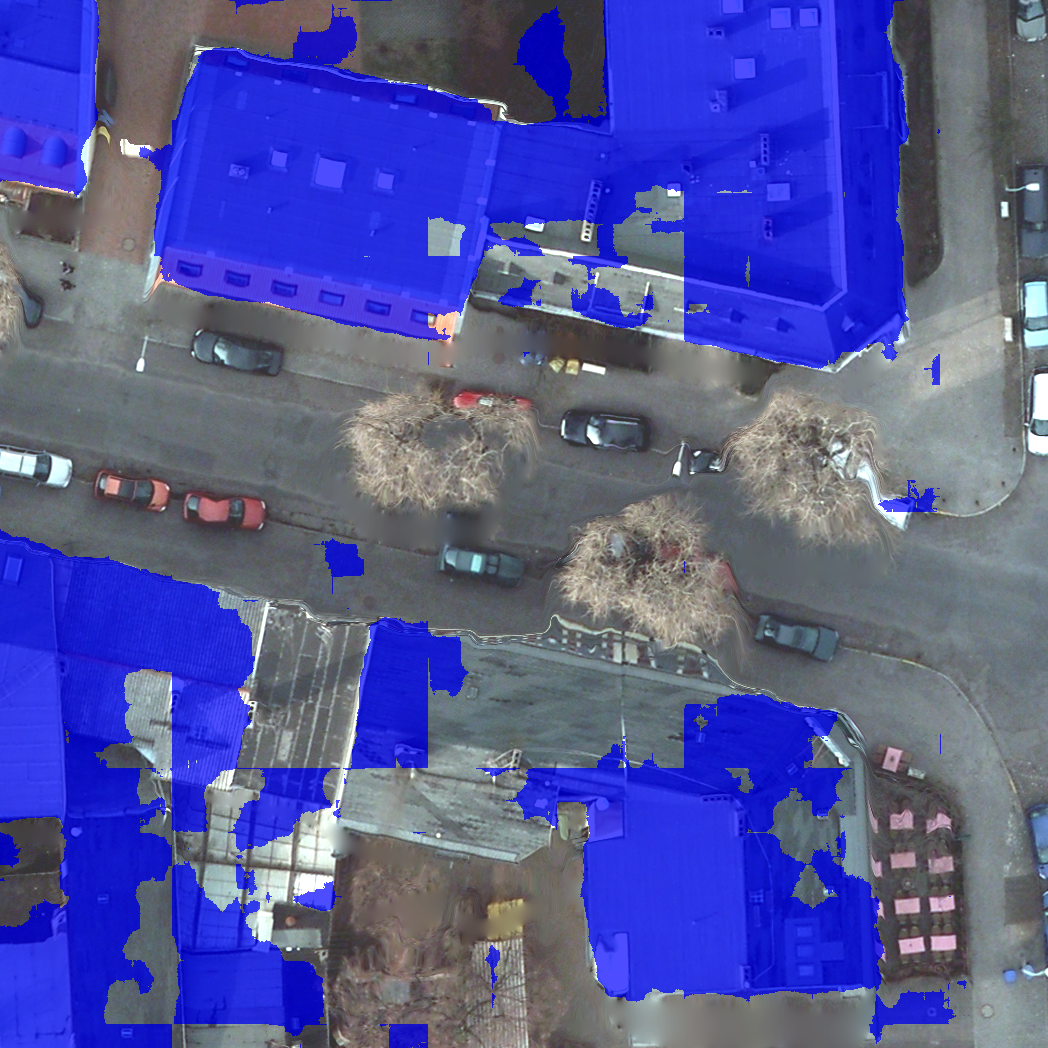}} &
    \subfloat[]{\includegraphics[width=0.45\textwidth]{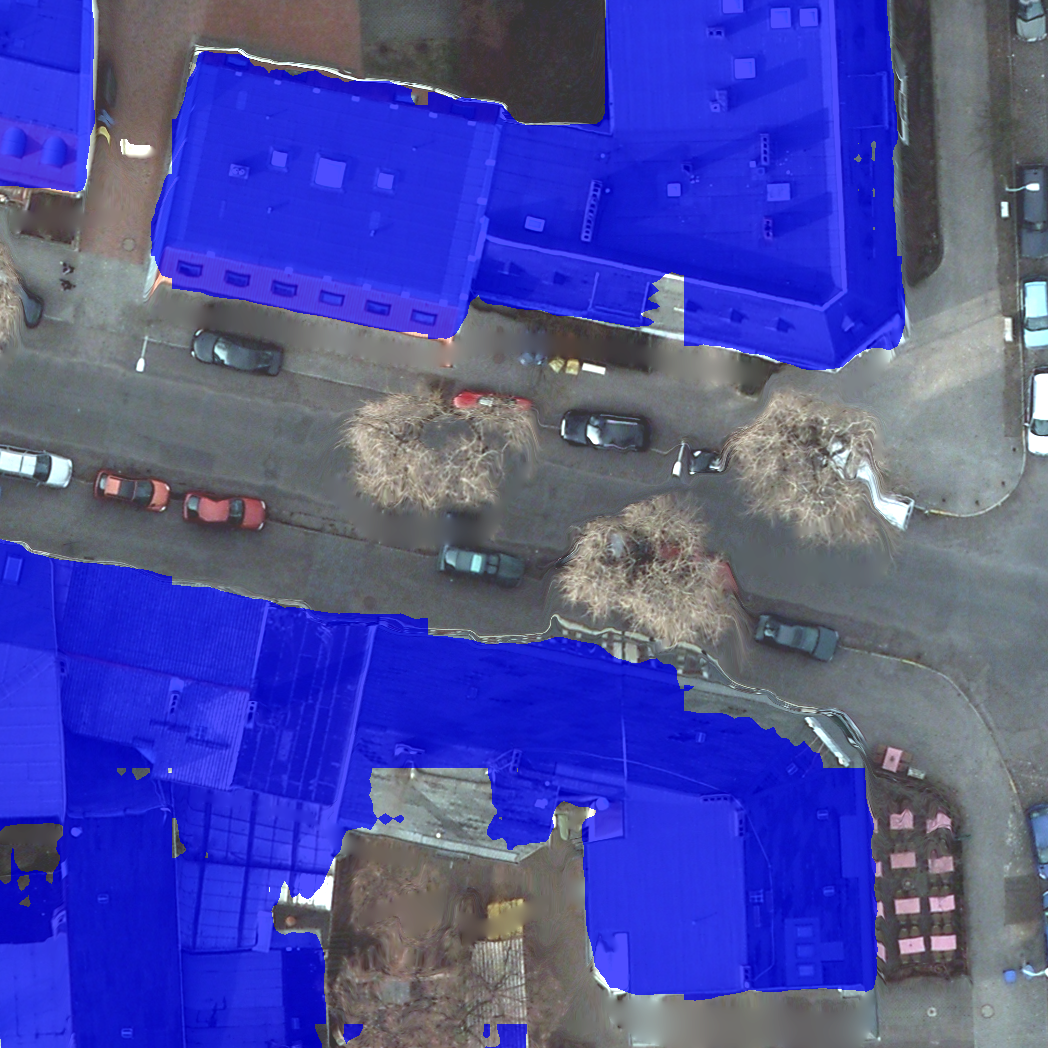}} \\

    \subfloat[]{\includegraphics[width=0.45\textwidth]{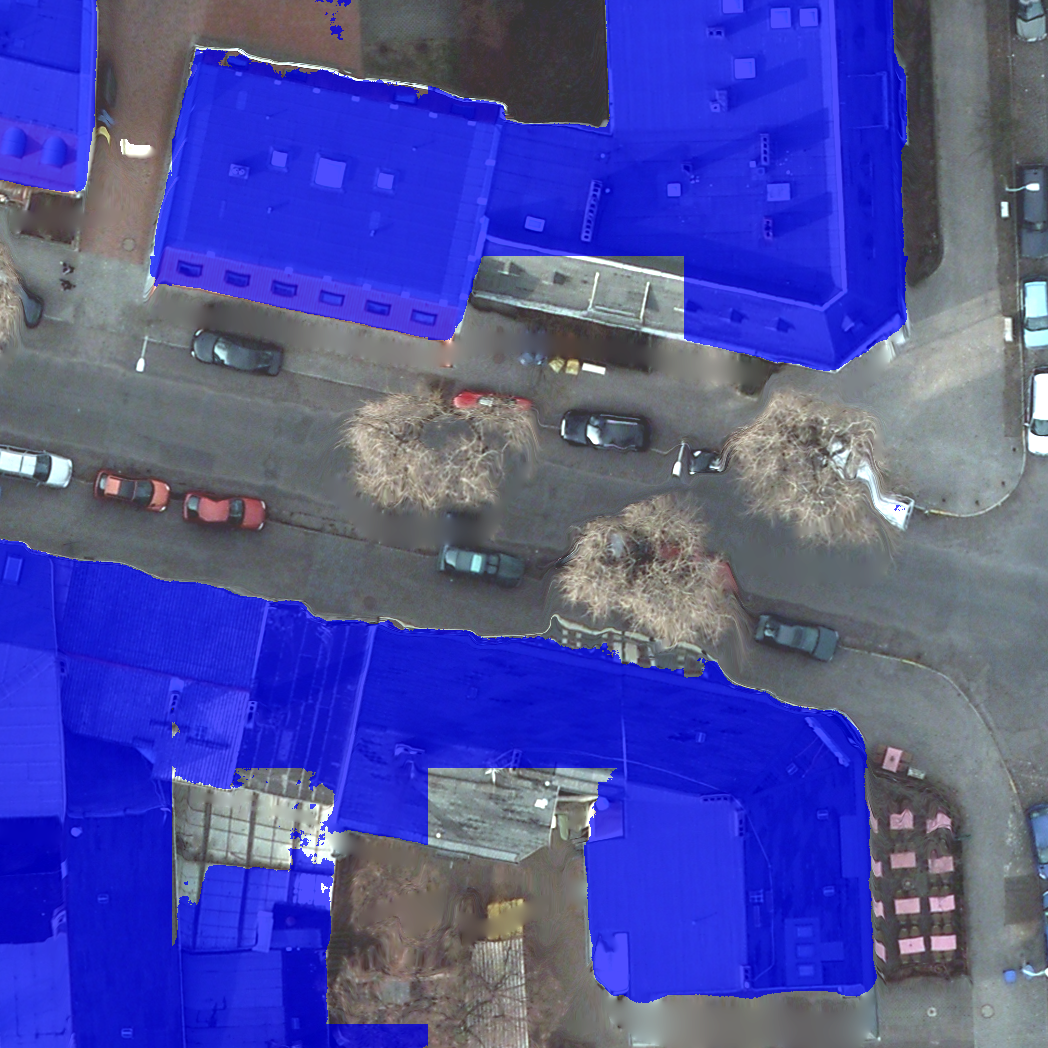}} &
    \subfloat[]{\includegraphics[width=0.45\textwidth]{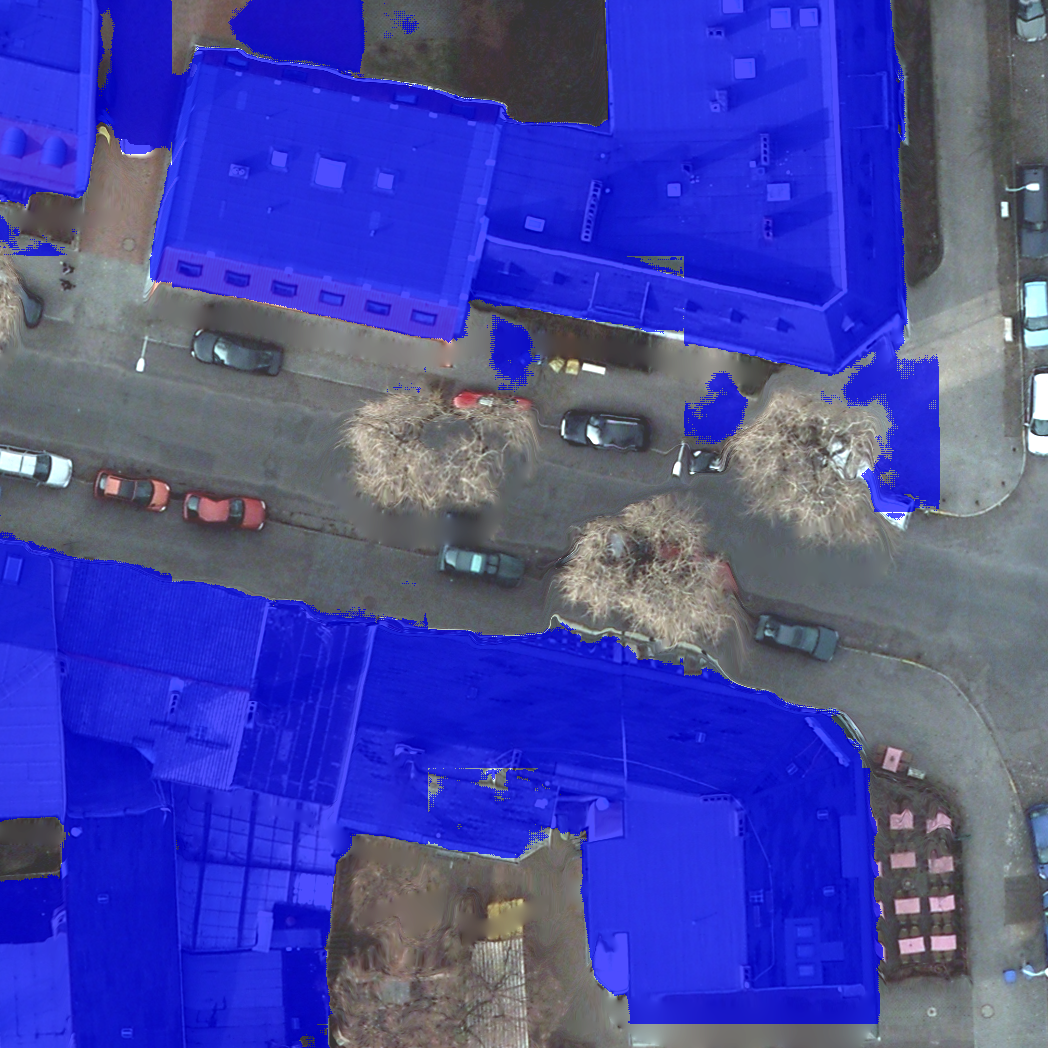}} \\

    \subfloat[]{\includegraphics[width=0.45\textwidth]{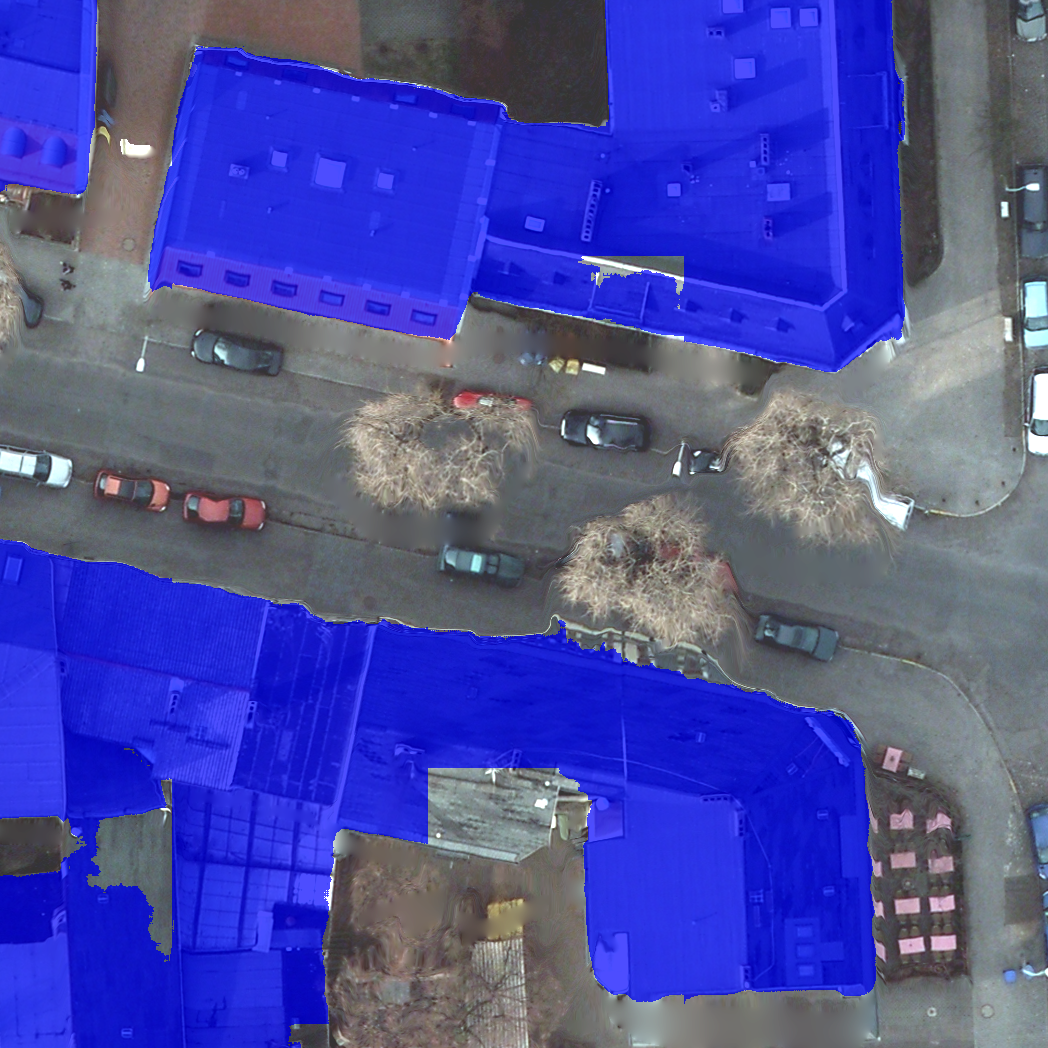}} &
    \subfloat[]{\includegraphics[width=0.45\textwidth]{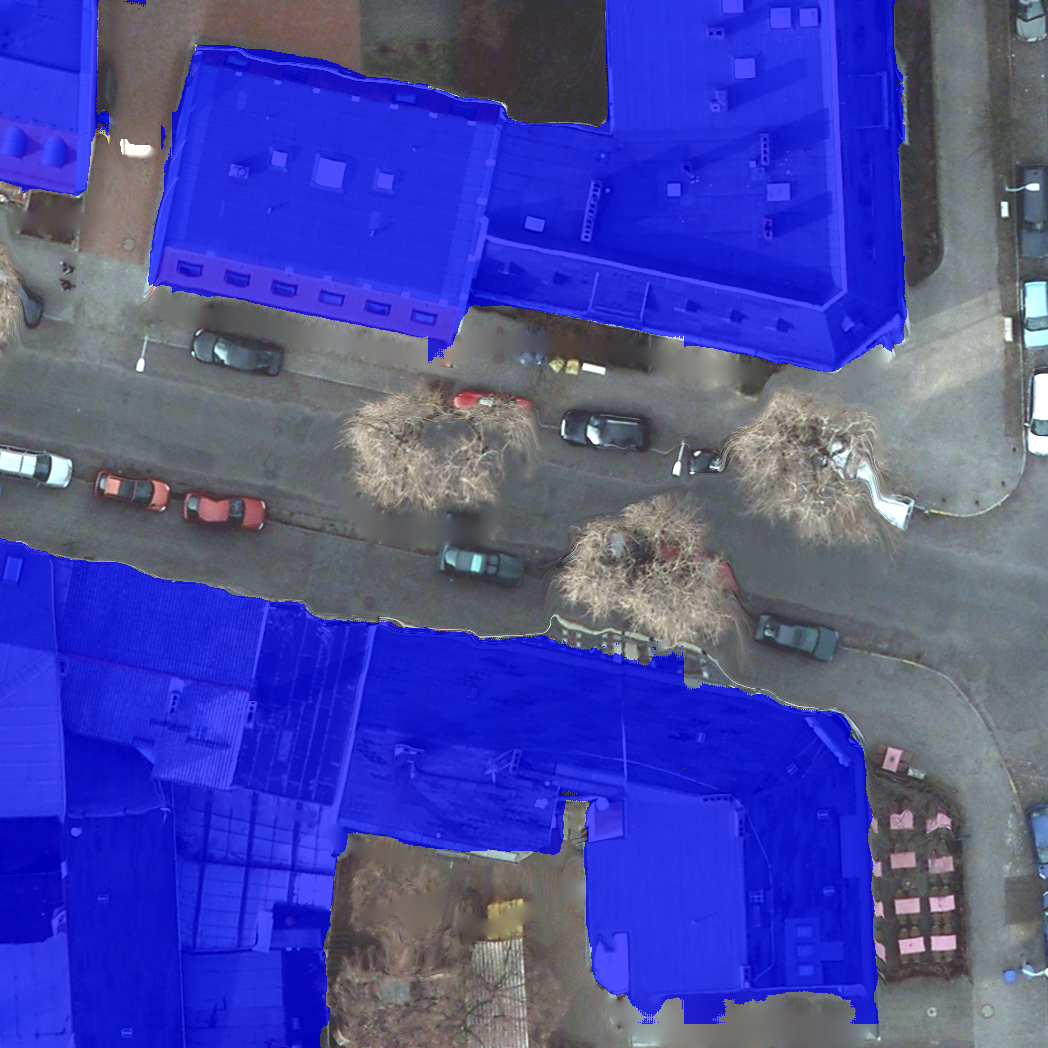}} \\

    \end{tabular}
    \caption{Visualized comparison of the predicted results on the ROI of ISPRS Potsdam dataset using different networks. (a) FCN-32; (b) FCN-16s; (c) ResNet-DUC; (d) E-Net; (e) SegNet; (f) U-Net; (g) FCN-8s; (h) CWGAN-GP; (i) FC-DenseNet; (j) DSFE-CRF; (k) DSFE-GCN; (l) DSFE-GGCN; (m) Ground truth; (n) Optical image.}
    \label{fig:comp_different_methods_isprs_small}
\end{figure}

\begin{figure}
  \ContinuedFloat
    \centering
    \begin{tabular}{cccc}
    \subfloat[]{\includegraphics[width=0.45\textwidth]{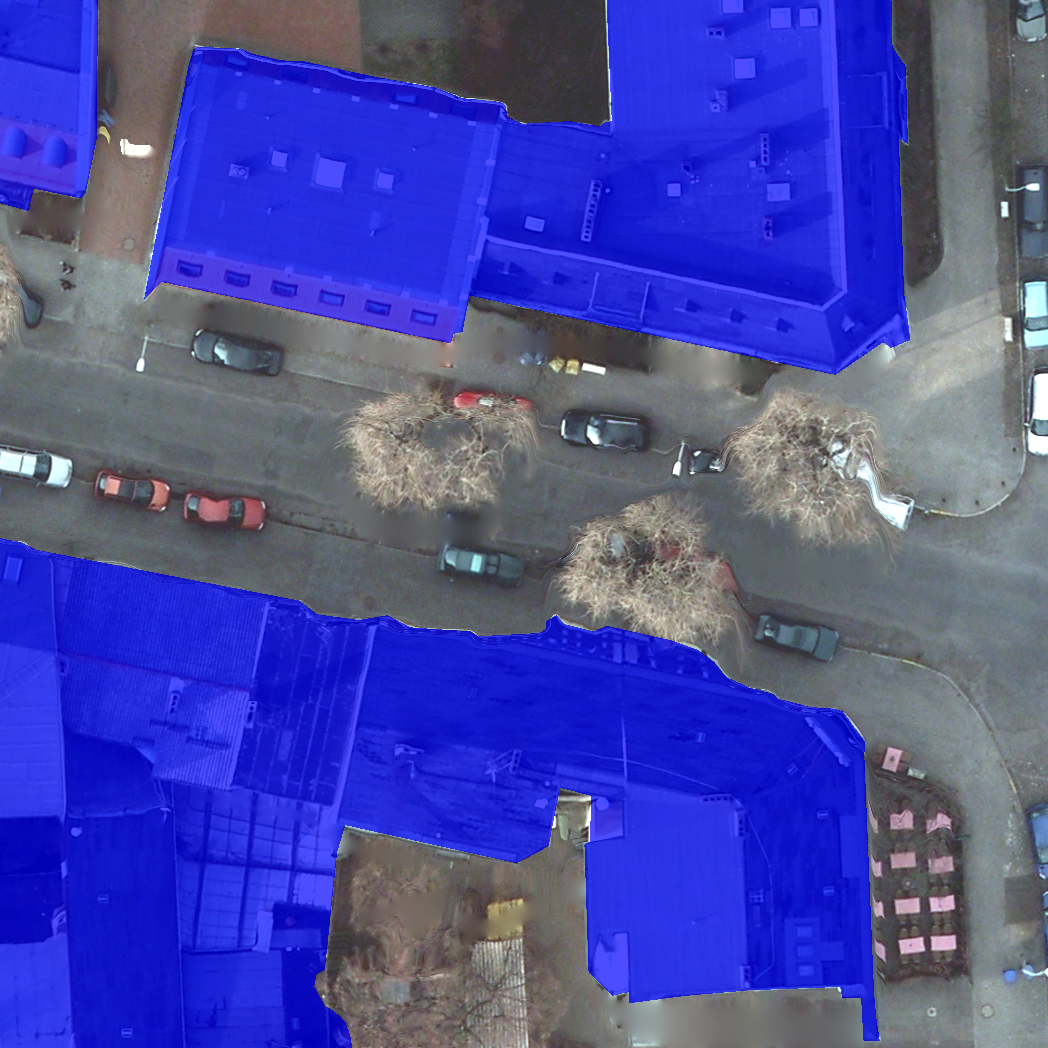}} &
    \subfloat[]{\includegraphics[width=0.45\textwidth]{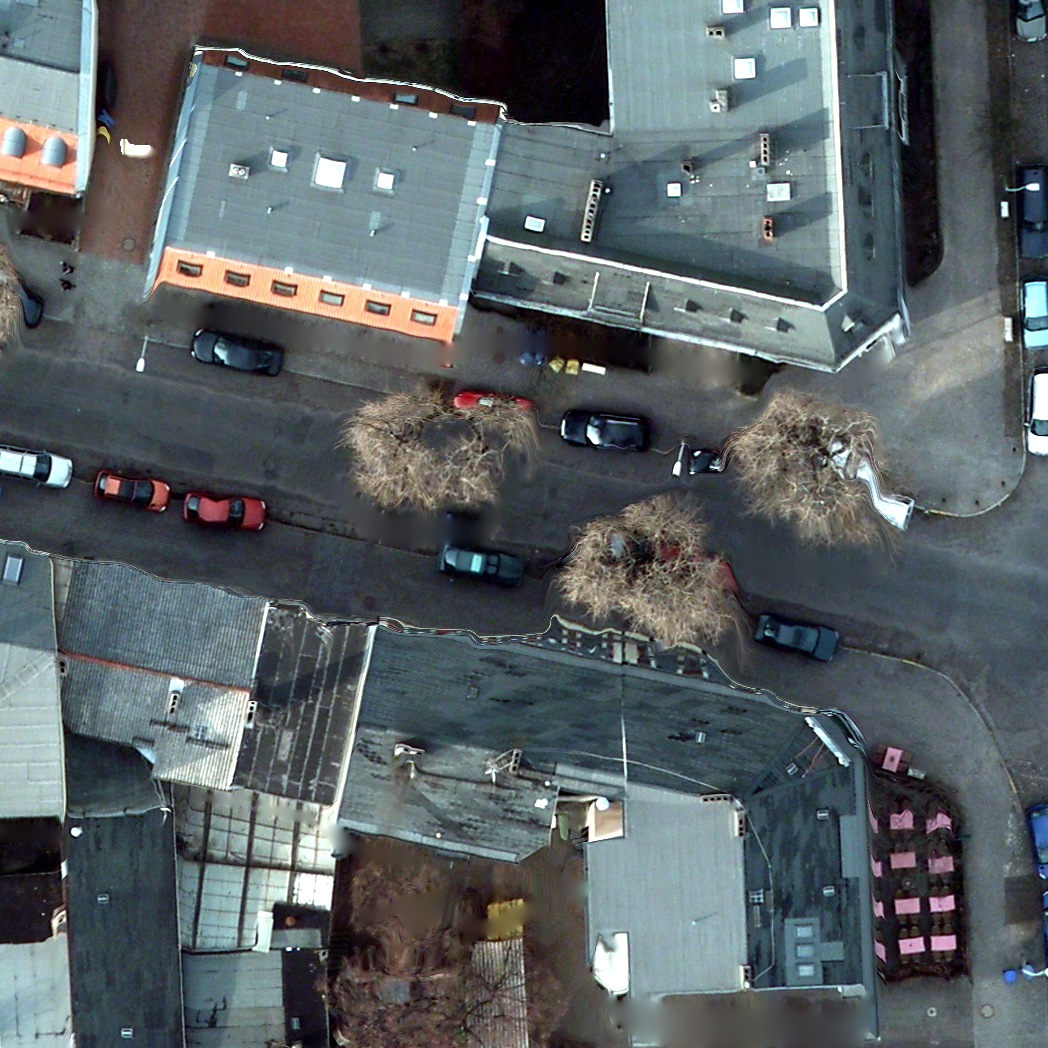}} & \\
    \end{tabular}
    \caption{Visualized comparison of the predicted results on the ROI of ISPRS Potsdam dataset using different networks. (a) FCN-32; (b) FCN-16s; (c) ResNet-DUC; (d) E-Net; (e) SegNet; (f) U-Net; (g) FCN-8s; (h) CWGAN-GP; (i) FC-DenseNet; (j) DSFE-CRF; (k) DSFE-GCN; (l) DSFE-GGCN; (m) Ground truth; (n) Optical image.}
    \label{fig:comp_different_methods_isprs_small}
\end{figure}

\begin{table}[ht]
\begin{center}
\begin{tabular}{cccc}
\toprule
\toprule
Methods  & \textbf{OA} & \textbf{F1} & \textbf{IoU} \\
\midrule
\rule{0pt}{2.5ex} FCN-32s  & 0.7371 & 0.6186 & 0.4478  \\
\rule{0pt}{2.5ex} FCN-16s  & 0.8247 & 0.7429 & 0.5910  \\
\rule{0pt}{2.5ex} ResNet-DUC  & 0.7475 & 0.6766 & 0.5051  \\
\rule{0pt}{2.5ex} E-Net  & 0.7711  & 0.7764 & 0.6110 \\
\rule{0pt}{2.5ex} SegNet & 0.8948  & 0.8511 & 0.7408 \\
\rule{0pt}{2.5ex} U-Net & 0.8892 & 0.8392  & 0.7229 \\
\rule{0pt}{2.5ex} FCN-8s & 0.8617 & 0.7986 & 0.6647 \\
\rule{0pt}{2.5ex} CWGAN-GP  & 0.8926 & 0.8504 & 0.7397 \\
\rule{0pt}{2.5ex} FC-DenseNet & 0.9186 & 0.9182 & 0.8789 \\

\rule{0pt}{2.5ex} DSFE-GCN & 0.9221 & 0.9375 & 0.9097 \\

\rule{0pt}{2.5ex} DSFE-GGCN & \textbf{0.9271} & \textbf{0.9422} & \textbf{0.9196} \\
\bottomrule
\bottomrule
\end{tabular}
\end{center}
\caption{Comparison between different deep convolutional neural networks and proposed framework on the ISPRS dataset}
\label{tab:comp_dcnn_isprs_results}
\end{table}

Table \ref{tab:comp_dcnn_isprs_results} summarizes the results of using different deep convolutional neural networks and the proposed framework on the ISPRS dataset. As can be seen the proposed DSFE-GGCN/DSFE-GCN framework contributes a significant improvement over the DCNNs. Moreover, compared to DSFE-GCN, DSFE-GGCN can effectively propagate the information in the short- and long-range, which leads to better results.

\section{Conclusion}
In this work, we develop a novel framework for semantic segmentation that combines the deep structured feature embedding and a graph convolutional network. Specifically, we propose using a gated graph convolutional network to improve the information propagation by using RNN with GCN. Our proposed framework outperforms the state-of-the-art methods for building footprint extraction. Although we have used building footprint extraction as the practical application, the proposed method can be generally applied to other binary or multi-label segmentation tasks, such as road extraction, settlement layer extraction, or semantic segmentation of very high resolution data in general. In addition, the proposed GCN network can work directly with unstructured data, such as point clouds and social media text messages.

\section{Acknowledgments}
This work is supported by the European Research Council (ERC) under the European Union's Horizon 2020 research and innovation programme (grant agreement no. ERC-2016-StG-714087, acronym: So2Sat, www.so2sat.eu), the Helmholtz Association under the framework of the Young Investigators Group ``SiPEO" (VH-NG-1018, www.sipeo.bgu.tum.de), Munich Aerospace e.V. Fakult{\"a}t f{\"u}r Luft- und Raumfahrt, and the Bavaria California Technology Center (Project: Large-Scale Problems in Earth Observation). The authors thank the Gauss Centre for Supercomputing (GCS) e.V. for funding this project by providing computing time on the GCS Supercomputer SuperMUC at the Leibniz Supercomputing Centre (LRZ) and on the supercomputer JURECA at Forschungszentrum J{\"u}lich. The authors thank Planet for providing the datasets.

\end{document}